\pdfoutput=1
\documentclass[letterpaper, 10 pt, conference]{ieeeconf}  

\IEEEoverridecommandlockouts                              

\usepackage{amsmath}
\usepackage{amssymb}
\usepackage{algpseudocode}
\usepackage[pdftex]{graphicx}
\usepackage{url}
\usepackage{siunitx}
\usepackage[percent]{overpic}
\usepackage[caption=false, font=footnotesize]{subfig}
\usepackage[export]{adjustbox}
\usepackage{upgreek}

\graphicspath{{figures/}}
\DeclareGraphicsExtensions{.pdf,.jpeg,.png,.jpg}

\newcommand{\optx}{\mathbf{x}}
\newcommand{\opty}{\mathbf{y}}
\newcommand{\opts}{\boldsymbol{\lambda}}
\newcommand{\optz}{\boldsymbol{\mu}}
\newcommand{\optsmat}{\boldsymbol{\Lambda}}
\newcommand{\optzmat}{\mathbf{M}}

\newcommand{\abbalphacoll}{\beta_\text{coll}}
\newcommand{\abbprocupdatenetwork}{\textsc{UpdateNetwork}}
\newcommand{\abbprocbuildtraj}{\textsc{BuildTraj}}
\newcommand{\abbviolcost}{c}
\newcommand{\abbreward}{r}
\newcommand{\abbdiscountedreward}{\widetilde{\abbreward}}
\newcommand{\abbeqviolcost}{\abbviolcost_\text{eq}}
\newcommand{\abbineqviolcost}{\abbviolcost_\text{in}}
\newcommand{\abbupdatedirrhs}{\mathbf{d}}
\newcommand{\abbconvexobject}{\mathcal{L}}
\newcommand{\abbclosestpoint}{\mathbf{p}}
\newcommand{\abbdistance}{d}
\newcommand{\abbdistancevel}{\dot{d}}
\newcommand{\abbdistanceunitvector}{\mathbf{n}}
\newcommand{\abbsecuritydistance}{\abbdistance_{m}}
\newcommand{\abbinfluencedistance}{\abbdistance_{M}}
\newcommand{\abbmassmatrix}{\mathbf{H}}
\newcommand{\abbbiasmatrix}{\mathbf{C}}
\newcommand{\abbbaseforcetorque}{\mathbf{F}_\text{base}}
\newcommand{\abbidentitymat}{\mathbf{I}}
\newcommand{\abbzeromat}{\mathbf{0}}
\newcommand{\abbenv}{\mathcal{E}}
\newcommand{\abbnetwork}{\mathcal{N}}
\newcommand{\abbnetworkup}{\abbnetwork_\text{UP}}
\newcommand{\abbnetworkcpc}{\abbnetwork_\text{CPC}}
\newcommand{\abboptlayer}{\mathcal{O}}

\newcommand{\abbdoconstrained}{constrained}
\newcommand{\abbgenpos}{\mathbf{q}}
\newcommand{\abbgenvel}{\dot{\abbgenpos}}
\newcommand{\abbgenacc}{\ddot{\abbgenpos}}
\newcommand{\abbstate}{\mathbf{s}}
\newcommand{\abbaction}{\mathbf{a}}
\newcommand{\abbprediction}{\widetilde{\mathbf{a}}}
\newcommand{\abbcorrection}{\mathbf{a}^{*}}
\newcommand{\abboptvar}{\mathbf{x}}
\newcommand{\abbobjmat}{\mathbf{P}}
\newcommand{\abbobjvec}{\mathbf{q}}
\newcommand{\abbeqmat}{\mathbf{A}}
\newcommand{\abbeqvec}{\mathbf{b}}
\newcommand{\abbineqmat}{\mathbf{G}}
\newcommand{\abbineqvec}{\mathbf{h}}
\newcommand{\abbineqmatcst}{\abbineqmat_\text{cst}}
\newcommand{\abbineqveccst}{\abbineqvec_\text{cst}}
\newcommand{\abbineqmatlin}{\abbineqmat_\text{aff}}
\newcommand{\abbineqveclin}{\abbineqvec_\text{aff}}
\newcommand{\abbineqmataux}{\abbineqmat_\text{aux}}
\newcommand{\abbineqvecaux}{\abbineqvec_\text{aux}}
\newcommand{\abbineqmatcond}{\abbineqmat_\text{cond}}
\newcommand{\abbineqveccond}{\abbineqvec_\text{cond}}
\newcommand{\abbineqmatbasesubset}{\abbineqmat_\text{sub}}
\newcommand{\abbineqvecbasesubset}{\abbineqvec_\text{sub}}

\newcommand{\abbineqvectest}{\abbineqvec_\text{test}}
\newcommand{\abbineqmatbase}{\abbineqmat_\text{base}}
\newcommand{\abbineqvecbase}{\abbineqvec_\text{base}}
\newcommand{\abbnoptvar}{n_\abboptvar}
\newcommand{\abbnineq}{n_\text{in}}
\newcommand{\abbneq}{n_\text{eq}}
\newcommand{\abbx}{\mathbf{x}}
\newcommand{\abby}{\mathbf{y}}
\newcommand{\abbz}{\mathbf{z}}
\newcommand{\abbjacobian}{\mathbf{J}}
\newcommand{\abbjointtorques}{\boldsymbol{\uptau}}
\newcommand{\abbjointpos}{\boldsymbol{\uptheta}}
\newcommand{\abbjointvel}{\dot{\abbjointpos}}
\newcommand{\abbjointacc}{\ddot{\abbjointpos}}
\newcommand{\abbendeffectorpos}{\mathbf{p}_\text{ee}}
\newcommand{\abbendeffectorposi}{\mathbf{p}_{\text{ee},i}}
\newcommand{\abbtargetpos}{\mathbf{p}_\text{t}}
\newcommand{\abbobstaclepos}{\mathbf{p}_\text{o}}
\newcommand{\abbdistancetargetee}{d_\text{T}}
\newcommand{\abbforearmpos}{\mathbf{p}_\text{fa}}
\newcommand{\abbselectionmatrixlin}{\mathbf{H}_{\abbjointpos}}
\newcommand{\abbselectionmatrixtest}{\mathbf{H}_{\abbdistance_l}}
\newcommand{\abbqpx}{\optx}
\newcommand{\abbqpy}{\opty}
\newcommand{\abbqpz}{\optz}
\newcommand{\abbqps}{\opts}
\newcommand{\abbqpzmat}{\optzmat}
\newcommand{\abbqpsmat}{\optsmat}

\usepackage{tikz}
\usetikzlibrary{backgrounds, shapes.geometric, arrows, fit, fadings}
\tikzstyle{state} = [rectangle, thick, rounded corners, draw=black, align=center, fill=red!10]
\tikzstyle{statevar} = [rectangle, thick, draw=black, align=center, fill=red!20]
\tikzstyle{neuralnetwork} = [rectangle, thick, rounded corners, draw=black, align=center, fill=blue!10]
\tikzstyle{reinforce} = [rectangle, thick, rounded corners, draw=black, align=center, fill=blue!20]
\tikzstyle{optlayer} = [rectangle, thick, rounded corners, draw=black, align=center, fill=yellow!10]
\tikzstyle{optlayerbuild} = [rectangle, thick, rounded corners, draw=black, align=center, fill=orange!10]
\tikzstyle{optlayerviol} = [rectangle, thick, rounded corners, draw=black, align=center, fill=black!10]
\tikzstyle{optlayerinit} = [rectangle, thick, rounded corners, draw=black, align=center, fill=green!10]
\tikzstyle{optlayeriter1} = [rectangle, thick, rounded corners, draw=black, align=center, fill=green!30]
\tikzstyle{optlayeritern} = [rectangle, thick, rounded corners, draw=black, align=center, fill=green!40]
\tikzstyle{optlayerselect} = [rectangle, thick, rounded corners, draw=black, align=center, fill=green!50]
\tikzstyle{optlayersum} = [circle, thick, draw=black, align=center]
\tikzstyle{optlayerdashed} = [rectangle, thick, dashed, rounded corners, draw=black, align=center, fill=yellow!10]

\tikzstyle{startstop} = [rectangle, rounded corners, minimum width=3cm, minimum height=1cm,text centered, draw=black, fill=red!10]
\tikzstyle{startstop_wider} = [rectangle, rounded corners, minimum width=3.3cm, minimum height=1cm,text centered, draw=black, fill=red!10]
\tikzstyle{startstop_noround} = [rectangle,text centered, fill=red!10]
\tikzstyle{io} = [trapezium, trapezium left angle=70, trapezium right angle=110, minimum width=3cm, minimum height=1cm, text centered, draw=black, fill=blue!10]
\tikzstyle{data} = [rectangle, rounded corners,, minimum width=3cm, minimum height=1cm, text centered, draw=black, fill=blue!10]
\tikzstyle{data_wider} = [rectangle, rounded corners,, minimum width=3.3cm, minimum height=1.8cm, text centered, draw=black, fill=blue!10]
\tikzstyle{data_small} = [ellipse, trapezium left angle=70, trapezium right angle=110, minimum width=1.5cm, minimum height=1cm, text centered, draw=black, fill=blue!10]
\tikzstyle{process} = [rectangle, minimum width=3cm, minimum height=1cm, text centered, draw=black, fill=yellow!10]
\tikzstyle{decision} = [diamond, minimum width=3cm, minimum height=1cm, text centered, draw=black, fill=green!10]
\tikzstyle{arrow} = [thick,->,>=stealth]
\tikzstyle{arrow_ends} = [thick,<->,>=stealth]
\tikzstyle{line} = [thick,-,>=stealth]
\tikzstyle{bigbox} = [draw=blue!50, thick, fill=blue!10, rounded corners, rectangle]
\tikzstyle{network} = [draw=blue!50, thick, rounded corners, rectangle]
\tikzstyle{kdn} = [rectangle, rounded corners, minimum width=3cm, minimum height=1cm, text centered, draw=black, fill=green!10]
\tikzstyle{rnn} = [draw=black, thick, fill=red!10, rounded corners, rectangle, text centered]
\tikzstyle{fcl} = [draw=black, thick, fill=yellow!10, rounded corners, rectangle, text centered]
\tikzstyle{features} = [draw=black, thick, fill=blue!10, rounded corners, rectangle, text centered]

\tikzfading[name=fade right,
  left color=transparent!0, right color=transparent!100]

\usetikzlibrary{calc,positioning,shapes.misc}

\tikzset{
 shrink inner sep/.code={
   \pgfkeysgetvalue{/pgf/inner xsep}{\currentinnerxsep}
   \pgfkeysgetvalue{/pgf/inner ysep}{\currentinnerysep}
   \pgfkeyssetvalue{/pgf/inner xsep}{\currentinnerxsep - 0.5\pgflinewidth}
   \pgfkeyssetvalue{/pgf/inner ysep}{\currentinnerysep - 0.5\pgflinewidth}
   }
}

\tikzset{horizontal shaded border/.style args={#1 and #2}{
    append after command={
       \pgfextra{%
          \begin{pgfinterruptpath}
                \path[rounded corners,left color=#1,right color=#2]
                ($(\tikzlastnode.south west)+(-\pgflinewidth,-\pgflinewidth)$) 
                rectangle
                ($(\tikzlastnode.north east)+(\pgflinewidth,\pgflinewidth)$);        
           \end{pgfinterruptpath}
        } 
    }
  },
  vertical shaded border/.style args={#1 and #2}{
    append after command={
       \pgfextra{%
          \begin{pgfinterruptpath}
                \path[rounded corners,top color=#1,bottom color=#2]
                ($(\tikzlastnode.south west)+(-\pgflinewidth,-\pgflinewidth)$) 
                rectangle
                ($(\tikzlastnode.north east)+(\pgflinewidth,\pgflinewidth)$);        
           \end{pgfinterruptpath}
        } 
    }
  }
}

\newcommand{\flowchartarrowhorizontalrightmiddlerighttextabove}[5]{
    \draw [arrow] ({#1}.east|-{#2}.west) -- node ({#5}) [align=center] {{#3} \\ {#4}} ({#2}.west);
}

\newcommand{\flowchartlineverticalcenterright}[3]{
    \draw [line] ({#1}.center) -- node ({#3}) {} ({#1}.south|-{#2}.west);
}
\newcommand{\flowchartarrowrighthorizontalright}[5]{
    \draw [arrow] ({#1}.east|-{#2}.center) -- node ({#5}) [align=center] {{#3} \\ {#4}} ({#2}.center);
}
\newcommand{\flowchartarrowrighthorizontalleft}[5]{
    \draw [arrow] ({#1}.east|-{#2}.center) -- node ({#5}) [align=center] {{#3} \\ {#4}} ({#2}.west);
}
\newcommand{\flowchartarrowcenterhorizontalright}[5]{
    \draw [arrow] ({#1}.south|-{#2}.center) -- node ({#5}) [align=center] {{#3} \\ {#4}} ({#2}.center);
}
\newcommand{\flowchartarrowwestcenterhorizontalright}[5]{
    \draw [arrow] ({#1}.west|-{#2}.center) -- node ({#5}) [align=center] {{#3} \\ {#4}} ({#2}.center);
}
\newcommand{\flowchartarrowcenterhorizontalrightmiddlerighttextabove}[5]{
    \draw [arrow] ({#1}.east|-{#2}.center) -- node ({#5}) [align=center] {{#3} \\ {#4}} ({#2}.center);
}
\newcommand{\flowchartarrowcenterverticalrightwest}[5]{
    \draw [line] ({#1}.center) -- node ({#3}) {} ({#1}.south|-{#2}.west);
    \draw [arrow] ({#1}.south|-{#2}.center) -- node ({#5}) [align=center] {{#3} \\ {#4}} ({#2}.west);
}

\newcommand{\flowchartarrowhorizontalvertical}[5]{
    \draw [line] ({#1}) -- node ({#3}) {} ({#1}-|{#2});
    \draw [arrow] ({#1}-|{#2}) -- node ({#5}) [align=center] {{#3} \\ {#4}} ({#2});
}
\newcommand{\flowchartarrowverticalhorizontal}[5]{
    \draw [arrow] ({#1}|-{#2}) -- node ({#5}) [align=center] {{#3} \\ {#4}} ({#2});
    \draw [line] ({#1}) -- node ({#3}) {} ({#1}|-{#2});
}
\newcommand{\flowchartarrowrighthorizontalleftalignleft}[5]{
    \draw [arrow] ({#1}.east) -- node ({#5}) [align=center] {{#3} \\ {#4}} ({#1}.east-|{#2}.west);
}
\newcommand{\flowchartarrowrighthorizontalrightalignleft}[5]{
    \draw [arrow] ({#1}.east) -- node ({#5}) [align=center] {{#3} \\ {#4}} ({#1}.east-|{#2}.east);
}

\newcommand{\flowchartoverview}{
    \begin{tikzpicture}
        \node (state_box) [state, minimum width=2.5cm, minimum height=3cm] {};
        \node (state_text) [align=center, yshift=1.2cm] at (state_box.center) {Environment};
        \node (state_robot_box) [statevar, below of=state_text, yshift=0.4cm, minimum width=2.1cm, minimum height=0.5cm] {Agent};
        \node (state_environment_box) [statevar, below of=state_robot_box, yshift=0.4cm, minimum width=2.1cm, minimum height=0.5cm] {World};
        \node (state_environment_box) [statevar, align=center, below of=state_environment_box, yshift=0.15cm, minimum width=2.1cm, minimum height=0.5cm] {Constraint \\ parameters};

        \node (neural_network_box) [neuralnetwork, right of=state_box, minimum width=2.0cm, minimum height=2.0cm, xshift=2cm, yshift=0.5cm] {};
        \node (neural_network_text) [align=center] at (neural_network_box.center) {Neural \\ network};

        \node (optlayer_box) [optlayer, right of=state_box, minimum width=2.0cm, minimum height=2.0cm, xshift=4.7cm, yshift=-0.5cm] {};
        \node (optlayer_text) [align=center] at (optlayer_box.center) {OptLayer};
        \node (optlayer_west_1third) [align=center, yshift=0.4cm] at (optlayer_box.west) {};
        \node (optlayer_west_2thirds) [align=center, yshift=-0.4cm] at (optlayer_box.west) {};
        \node (optlayer_east_1third) [align=center, yshift=0.4cm] at (optlayer_box.east) {};
        \node (optlayer_east_2thirds) [align=center, yshift=-0.4cm] at (optlayer_box.east) {};

        \node (output_optlayer) [right of=optlayer_east_2thirds, minimum width=0.1cm, minimum height=0.1cm, xshift=-0.3cm, yshift=0cm] {};
        \node (output_viol) [right of=optlayer_east_1third, minimum width=0.1cm, minimum height=0.1cm, xshift=-0.3cm, yshift=0cm] {};
        \node (output_value) [above of=output_viol, minimum width=0.1cm, minimum height=0.1cm, xshift=0.0cm, yshift=0.5cm] {};
        \node (output_raw) [above of=output_viol, minimum width=0.1cm, minimum height=0.1cm, xshift=0.0cm, yshift=-0.2cm] {};

        \flowchartarrowhorizontalrightmiddlerighttextabove{state_box}{neural_network_box}{$\mathbf{s}$}{}{midarrow_state}
        \flowchartlineverticalcenterright{midarrow_state}{optlayer_west_2thirds}{}
        \flowchartarrowcenterhorizontalright{midarrow_state}{optlayer_west_2thirds}{}{}{secondarrow_state}
        \flowchartarrowcenterhorizontalrightmiddlerighttextabove{neural_network_box}{optlayer_west_1third}{$\widetilde{\mathbf{a}}$}{}{midarrow_raw}
        \node (midarrow_raw_shift) [align=center, xshift=0.15cm] at (midarrow_raw.center) {};
        \flowchartlineverticalcenterright{midarrow_raw_shift}{output_raw}{}
        \flowchartarrowcenterhorizontalright{midarrow_raw_shift}{output_raw}{}{}{endarrow_output_raw}
        \node (endarrow_output_raw_text) [align=center, xshift=1.05cm] at (endarrow_output_raw.center) {$\widetilde{\mathbf{a}}$ \\ };
        \node (secondarrow_state_text) [align=center, xshift=1.2cm] at (secondarrow_state.center) {$\mathbf{s}$ \\ };
        \flowchartarrowcenterhorizontalrightmiddlerighttextabove{optlayer_box}{output_optlayer}{$\mathbf{a}^*$}{}{}
        \flowchartarrowcenterhorizontalrightmiddlerighttextabove{optlayer_box}{output_viol}{}{}{viol_arrow}
        \node (viol_text) [align=center, xshift=-0.05cm] at (viol_arrow.center) {${c}$ \\ };
        \flowchartarrowrighthorizontalright{neural_network_box}{output_value}{}{}{midarrow_value}
        \node (endarrow_value_text) [align=center, xshift=1.28cm] at (midarrow_value.center) {${v}$ \\ };

    \end{tikzpicture}
}

\newcommand{\flowchartoptlayer}{
    \begin{tikzpicture}
        \node (optlayer_box) [optlayerdashed, minimum width=14.6cm, xshift=6.00cm, minimum height=4.7cm, yshift=0.55cm] {};
        \node (optlayer_text) [align=center, yshift=1.90cm] at (optlayer_box.center) {
                OptLayer:
                $
                \min\limits_{\mathbf{x}} {
                    \frac{1}{2}\mathbf{x}^T\mathbf{P}\mathbf{x}
                    + \mathbf{q}^T\mathbf{x}
                }
                \,\,
                \text{such that}
                \,\,
                \mathbf{G}\mathbf{x} \leq \mathbf{h}
                \,\,
                \text{and}
                \,\,
                \mathbf{A}\mathbf{x} = \mathbf{b}
                $
        };

        \node (build_matrices_box) [optlayerbuild, minimum width=1.7cm, minimum height=3.0cm] {};
        \node (build_matrices_text) [align=center] at (build_matrices_box.center) {Build \\ constraint \\ matrices};
        \node (build_matrices_box_west_1third) [align=center, yshift=0.7cm] at (build_matrices_box.west) {};
        \node (build_matrices_box_west_2thirds) [align=center, yshift=-0.7cm] at (build_matrices_box.west) {};

        \node (input_action) [left of=build_matrices_box, minimum width=0.01cm, minimum height=0.1cm, xshift=-0.6cm, yshift=0cm] {};
        \node (input_state) [left of=build_matrices_box, minimum width=0.01cm, minimum height=0.1cm, xshift=-0.6cm, yshift=0cm] {};
        \flowchartarrowwestcenterhorizontalright{input_action}{build_matrices_box_west_1third}{}{}{arrow_in_action}
        \flowchartarrowwestcenterhorizontalright{input_state}{build_matrices_box_west_2thirds}{}{}{arrow_in_state}
        \node (arrow_in_action_split) [align=center, xshift=0.18cm] at (arrow_in_action.center) {};
        \node (arrow_in_action_text) [align=center, xshift=-0.25cm] at (arrow_in_action.center) {$\widetilde{\mathbf{a}}$ \\ };
        \node (arrow_in_state_text) [align=center, xshift=-0.25cm] at (arrow_in_state.center) {$\mathbf{s}$ \\ };

        \node (init_opt_box) [optlayerinit, right of=state_box, minimum width=1.2cm, minimum height=2.0cm, xshift=1.5cm, yshift=-0.5cm] {};
        \node (init_opt_text) [align=center] at (init_opt_box.center) {Init. \\ optim. \\ var.};
        \flowchartarrowhorizontalrightmiddlerighttextabove{build_matrices_box}{init_opt_box}{$\mathbf{P}, \! \mathbf{q}, \! \mathbf{G}$}{$\mathbf{h}, \! \mathbf{A}, \! \mathbf{b}$}{midarrow_init_var}

        \node (violation_box) [optlayerviol, right of=state_box, minimum width=3.2cm, minimum height=1.0cm, xshift=2.4cm, yshift=1.6cm] {};
        \node (violation_box_west) [align=center] at (violation_box.west) {};
        \node (violation_box_west_1third) [align=center, yshift=0.2cm] at (violation_box.west) {};
        \node (violation_box_west_2thirds) [align=center, yshift=-0.2cm] at (violation_box.west) {};
        \node (violation_text) [align=center] at (violation_box.center) {Compute constraint \\ violation cost};
        \flowchartlineverticalcenterright{midarrow_init_var}{violation_box_west_2thirds}{}
        \flowchartarrowcenterhorizontalright{midarrow_init_var}{violation_box_west_2thirds}{}{}{}
        \flowchartarrowverticalhorizontal{arrow_in_action_split.center}{violation_box_west_1third.center}{}{}{}

        \node (iter1_box) [optlayeriter1, right of=init_opt_box, minimum width=1.2cm, minimum height=1.5cm, xshift=0.6cm, yshift=0.25cm] {Optim. \\ iter. 1};
        \node (iter1_sum_box) [optlayersum, right of=iter1_box, radius=0.1cm, xshift=-0.23cm, yshift=-1.0cm] {};
        \node (iter1_sum_text) [align=center] at (iter1_sum_box.center) {$+$};
        \flowchartarrowrighthorizontalleft{init_opt_box}{iter1_sum_box}{}{}{iter1_arrow_in}
        \node (iter1_arrow_in_text) [align=center, xshift=0.07cm, yshift=-0.25cm] at (iter1_arrow_in.center) {\footnotesize{$\optx_0, \opty_0, \opts_0, \optz_0$}};
        \node (iter1_arrow_in_split) [align=center, xshift=-0.61cm] at (iter1_arrow_in.center) {};
        \flowchartarrowcenterverticalrightwest{iter1_arrow_in_split}{iter1_box}{}{}{}
        \flowchartarrowhorizontalvertical{iter1_box.east}{iter1_sum_box.north}{}{}{}
        \node (iter1_delta) [align=center, xshift=0.55cm, yshift=-0.00cm] at (iter1_box.east) {\footnotesize{$\Delta\optx_0$} \\ \footnotesize{$\Delta\opty_0$} \\ \footnotesize{$\Delta\opts_0$} \\ \footnotesize{$\Delta\optz_0$}};
        \node (iter1_out_arrow_end) [right of=iter1_sum_box, xshift=1.0cm] {$\cdots$};
        \flowchartarrowrighthorizontalleft{iter1_sum_box}{iter1_out_arrow_end}{}{}{iter1_arrow_out}
        \node (iter1_arrow_out_text) [align=center, xshift=0.10cm, yshift=-0.25cm] at (iter1_arrow_out.center) {\footnotesize{$\optx_1, \opty_1, \opts_1, \optz_1$}};

        \node (itern_box) [optlayeritern, right of=init_opt_box, minimum width=1.2cm, minimum height=1.5cm, xshift=5.4cm, yshift=0.25cm] {Optim. \\ iter. $n$};
        \node (itern_sum_box) [optlayersum, right of=itern_box, radius=0.1cm, xshift=-0.23cm, yshift=-1.0cm] {};
        \node (itern_sum_text) [align=center] at (itern_sum_box.center) {$+$};
        \flowchartarrowrighthorizontalleft{iter1_out_arrow_end}{itern_sum_box}{}{}{itern_arrow_in}
        \node (itern_arrow_in_text) [align=center, xshift=0.05cm, yshift=-0.25cm] at (itern_arrow_in.center) {\footnotesize{$\optx_{n\!-\!1}, \opty_{n\!-\!1}, \opts_{n\!-\!1}, \optz_{n\!-\!1}$}};
        \node (itern_arrow_in_split) [align=center, xshift=-0.29cm] at (itern_arrow_in.center) {};
        \flowchartarrowcenterverticalrightwest{itern_arrow_in_split}{itern_box}{}{}{}
        \flowchartarrowhorizontalvertical{itern_box.east}{itern_sum_box.north}{}{}{}
        \node (itern_delta) [align=center, xshift=0.70cm, yshift=-0.00cm] at (itern_box.east) {\footnotesize{${\Delta\optx_{n\!-\!1}}$} \\ \footnotesize{${\Delta\opty_{n\!-\!1}}$} \\ \footnotesize{$\Delta\opts_{n\!-\!1}$} \\ \footnotesize{$\Delta\optz_{n\!-\!1}$}};
        \node (itern_out_arrow_end) [right of=itern_sum_box, xshift=1.1cm] {$\cdots$};

        \node (itern_select_box) [optlayerselect, right of=init_opt_box, minimum width=1.2cm, minimum height=2.0cm, xshift=9.0cm, yshift=0.00cm] {Select \\ $\mathbf{a}^* \! = \! \optx_n$};
        \flowchartarrowrighthorizontalleftalignleft{itern_sum_box}{itern_select_box}{}{}{itern_select_in}
        \node (itern_select_in_text) [align=center, xshift=0.0cm, yshift=-0.25cm] at (itern_select_in.center) {\footnotesize{$\optx_n, \opty_n, \opts_n, \optz_n$}};

        \node (arrow_matrices_start) [align=center, yshift=1.3cm] at (midarrow_init_var.center) {};
        \flowchartarrowhorizontalvertical{arrow_matrices_start.center}{iter1_box.north}{}{}{}
        \flowchartarrowrighthorizontalleftalignleft{arrow_matrices_start}{iter1_out_arrow_end}{}{}{}
        \node (arrow_matrices_dots) [right of=arrow_matrices_start, xshift=4.45cm] {$\cdots$};
        \flowchartarrowhorizontalvertical{arrow_matrices_dots.east}{itern_box.north}{}{}{}
        \node (arrow_matrices_itern) [above of=itern_box, xshift=-0.80cm, yshift=0.3cm] {$\mathbf{P}, \mathbf{q}, \mathbf{G}, \mathbf{h}, \mathbf{A}, \mathbf{b}$};

        \node (out_virtual) [right of=itern_select_box, minimum width=0.0cm, minimum height=2.0cm, xshift=0.10cm, yshift=0.00cm] {};
        \flowchartarrowrighthorizontalrightalignleft{itern_select_box}{out_virtual}{}{}{midarrow_out_action}
        \node (endarrow_out_action_text) [align=center, xshift=0.10cm] at (midarrow_out_action.center) {$\mathbf{a}^*$ \\ };
        \flowchartarrowrighthorizontalrightalignleft{violation_box}{out_virtual}{}{}{midarrow_out_viol}
        \node (endarrow_out_viol_text) [align=center, xshift=4.15cm] at (midarrow_out_viol.center) {$c$ \\ };

    \end{tikzpicture}
}

\title{\LARGE \bf
    OptLayer - Practical Constrained Optimization for \\
    Deep Reinforcement Learning in the Real World
}

\author{Tu-Hoa Pham$^{1}$, Giovanni De Magistris$^{1}$ and Ryuki Tachibana$^{1}$
\thanks{
    $^{1}$IBM Research AI, Tokyo, Japan.
}%
}

\begin{document}

\maketitle
\thispagestyle{empty}
\pagestyle{empty}

\begin{abstract}

    While deep reinforcement learning techniques have recently produced
    considerable achievements on many decision-making problems,
    their use in robotics has largely been limited to simulated worlds
    or restricted motions,
    since unconstrained trial-and-error interactions in the real world
    can have undesirable
    consequences for the robot or its environment.
    To overcome such limitations,
    we propose a novel reinforcement learning architecture, OptLayer,
    that takes as inputs possibly unsafe actions predicted by a neural network
    and outputs the closest actions that satisfy chosen constraints.
    While
    learning control policies
    often
    requires
    carefully crafted rewards and penalties
    while exploring the range of possible actions,
    OptLayer ensures that only safe actions are actually executed
    and unsafe predictions are penalized during training.
    We demonstrate the effectiveness of our approach on robot reaching tasks,
    both simulated and in the real world.

\end{abstract}

\section{Introduction}

Over the recent years,
the rise of deep neural network architectures in multiple fields of science
and engineering has come together with
considerable achievements for
deep reinforcement learning techniques,
sometimes outperforming humans on high-dimensional problems
such as Atari games~\cite{nature:mnih:2015} and Go~\cite{nature:silver:2016}.
The robotics field has also greatly benefitted from these advances,
yielding impressive results on tasks otherwise difficult to model explicitly,
e.g.,
manipulation from visual inputs~\cite{jmlr:levine:2016}
and
robust locomotion on challenging terrains~\cite{arxiv:heess:2017}.
While visual understanding neural network pipelines can often be trained
in an end-to-end fashion
(e.g., inferring object poses directly from raw pixel information),
robot motion is typically initialized in the real world from
(partially) supervised learning,
which supposes the availability of ground-truth data or expert demonstrations.
In contrast,
the discovery of robot motion behaviours from scratch
is often limited to simulation
due to
time (to learn basic task features),
safety (e.g., collisions),
and other experimental constraints
(e.g., resetting knocked-down obstacles to a chosen state).
However, transferring control policies learned in simulation
to reality remains a challenge,
in particular due to model uncertainties (kinodynamics)~\cite{arxiv:christiano:2016}
and variability between simulated and real observations~\cite{iros:tobin:2017}.
Enabling deep reinforcement learning in the real world
is thus crucial for tasks that cannot be learned in simulation
and for which expert policies are not known.

\begin{figure}[!t]
    \centering
    \subfloat[Unconstrained action predictions can be dangerous to run on a real robot.] {
        \label{fig:simplereach:un}
        \includegraphics[height=0.21\columnwidth]{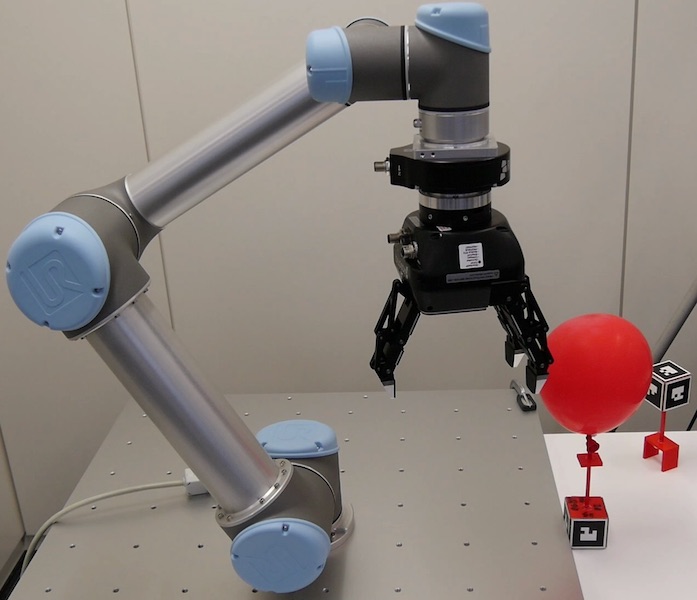}
        \includegraphics[height=0.21\columnwidth]{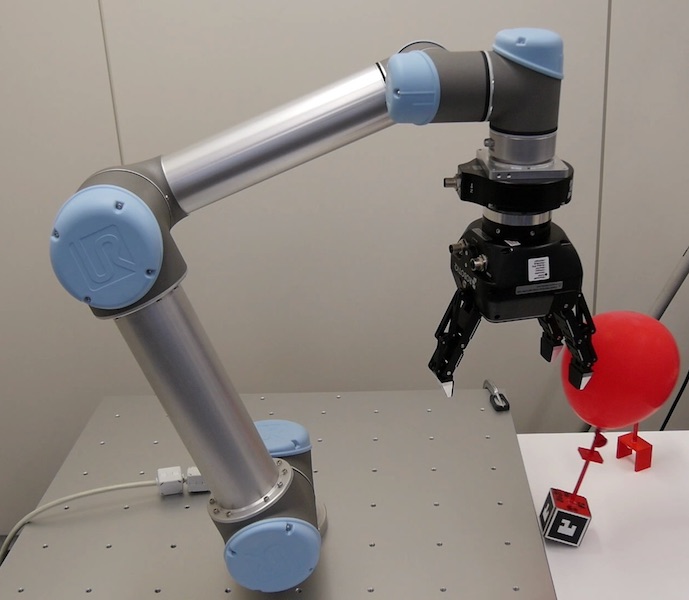}
        \includegraphics[height=0.21\columnwidth]{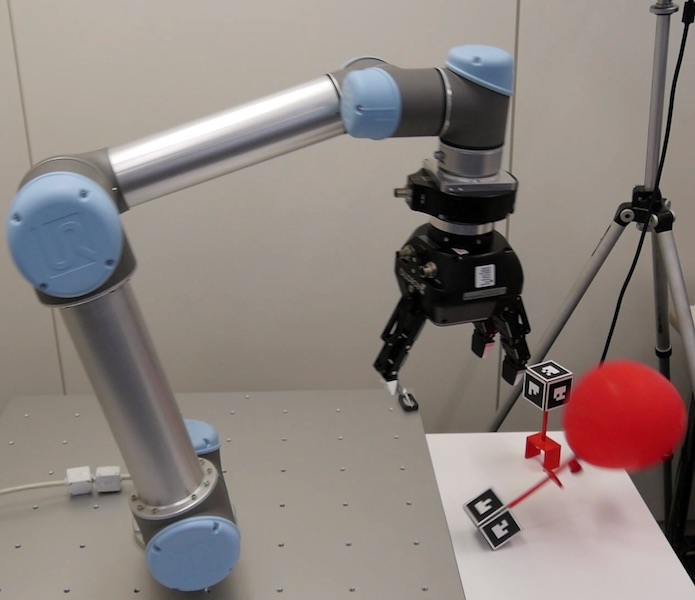}
        \includegraphics[height=0.21\columnwidth]{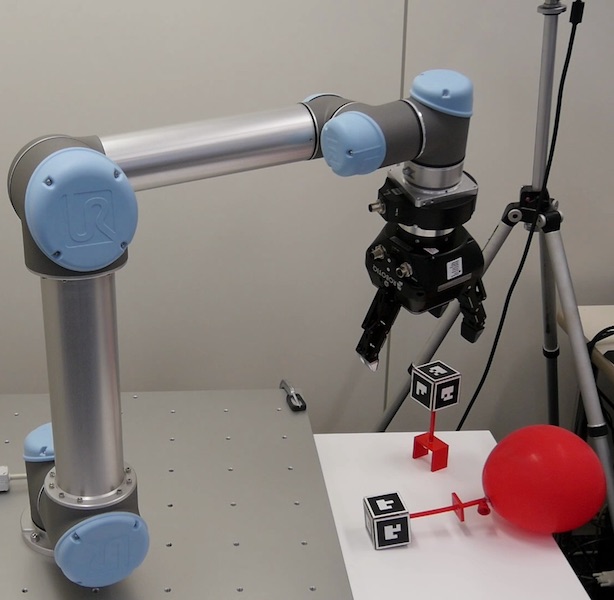}
    } \\
    \subfloat[Explicit constraints can be incorporated to guarantee safe robot actions.] {
        \label{fig:simplereach:co}
        \includegraphics[height=0.21\columnwidth]{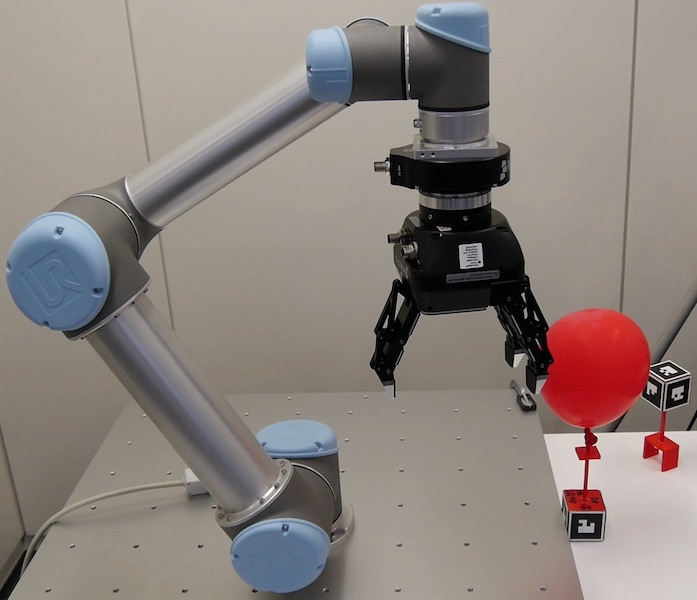}
        \includegraphics[height=0.21\columnwidth]{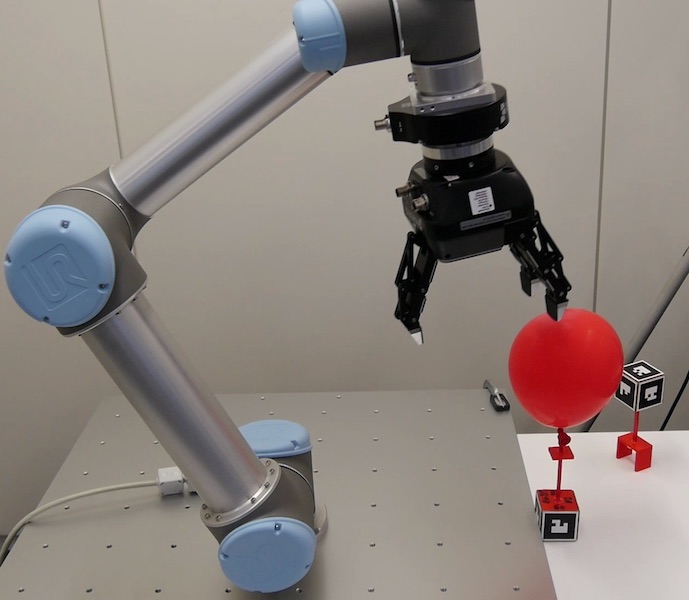}
        \includegraphics[height=0.21\columnwidth]{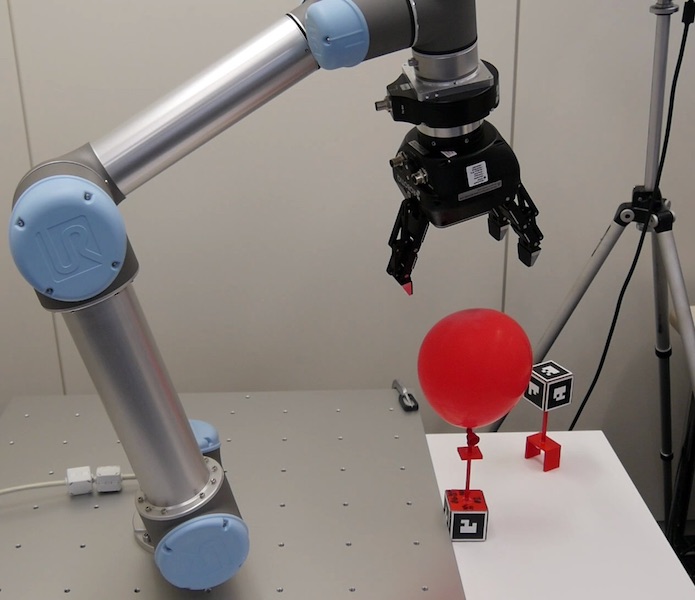}
        \includegraphics[height=0.21\columnwidth]{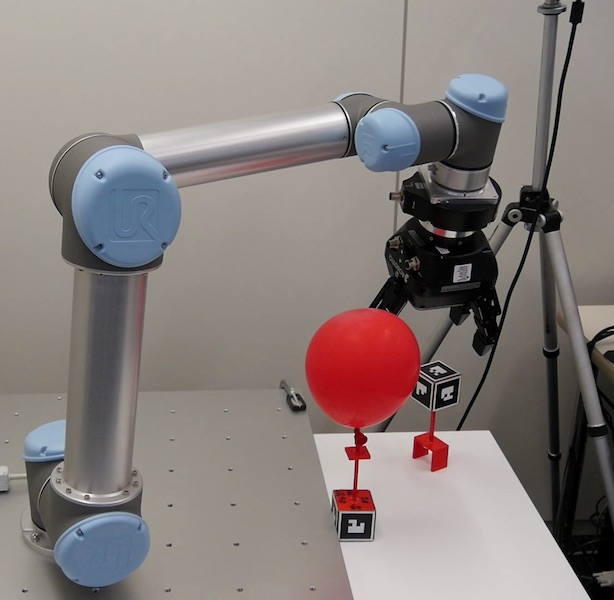}
    }
    \caption{
        3D reaching with obstacle avoidance using
        reinforcement learning.
    }
    \label{fig:simplereach}
\end{figure}

In this paper,
we propose a practical method to constrain neural network predictions
to lie within a domain defined by safety constraints (see Fig.~\ref{fig:simplereach}).
Our work capitalizes on
the state of the art in multiple areas of
robotic control, optimization and reinforcement learning
(Section~\ref{sec:literature_review}).
We summarize the necessary technical background and important challenges
for real-world reinforcement learning (Section~\ref{sec:problem_statement}),
in particular a strong dependency on the chosen reward structure.
\begin{itemize}
    \item We augment a neural network architecture
        with a constrained optimization layer,
        OptLayer,
        that enforces arbitrary constraints
        on the predicted robot actions
        (Section~\ref{sec:optlayer}).
        OptLayer is fully differentiable,
        enabling future
        end-to-end learning under safety constraints.
    \item We
        tackle the problem of
        safe reinforcement learning
        using OptLayer
        and propose a reward strategy
        that is simple to implement and readily compatible
        with existing policy optimization techniques
        (Section~\ref{sec:constrained_reinforcement_learning}).
    \item
        Our approach makes exploration during training more efficient
        while always satisfying safety constraints.
        We demonstrate its effectiveness
        by learning 3D reaching with collision avoidance
        on an industrial manipulator,
        on tasks of increasing difficulty,
        both simulated and in the real world
        (Section~\ref{sec:experiments}).
\end{itemize}
Finally, we discuss challenges we encountered,
current limitations
and future extensions of our work (Section~\ref{sec:discussion}).

\section{Related Work}
\label{sec:literature_review}

Robot motion is traditionally achieved by solving a multi-parameter optimization problem
yielding a sequence of states achieving a given task
(e.g., track a given end effector trajectory)
while respecting constraints motivated by safety
and performance (e.g., joint limits).
When constraints and task objectives are respectively linear and quadratic in the robot variables,
these problems can be formulated and solved as quadratic programs (QP)~\cite{oe:mattingley:2012}.
Still, the resulting control policy can be unsatisfactory due to
uncertainties on the robot kinodynamic model and sensor feedback
(e.g., during locomotion, errors on center of mass coordinates can invalidate robot stability constraints),
the chosen constraints being too conservative (making the motion suboptimal),
or the task dynamics being non-trivial to model (e.g., finding the optimal trajectory to follow).

Reinforcement learning (RL) techniques have demonstrated impressive results in this context.
In~\cite{nature:mnih:2015}, convolutional neural networks (CNN) were trained to play Atari games using only
pixels and game score as inputs, achieving human-level performance on multiple games without need for
expert demonstrations or feature engineering.
A guided policy search (GPS) method was proposed by~\cite{jmlr:levine:2016}
to predict motor torques for a PR2 robot from image and joint state inputs only,
training the policy in a data-efficient manner using supervised learning (SL)
and local controllers trained by RL.
An overview and benchmark of modern RL algorithms for continuous control
(e.g., for robotic applications)
was presented in~\cite{icml:duan:2016},
along with valuable reference implementations released to the community.
We thus chose to build our work upon the top-performing benchmarked algorithm,
Trust-Region Policy Optimization (TRPO)~\cite{icml:schulman:2015},
which we detail further in Section~\ref{sec:trpo}.
TRPO was notably used recently to discover dynamic locomotion behaviours
such as crouching and jumping
on simulated humanoids~\cite{arxiv:heess:2017}.

Applied in the real world, RL techniques require special attention to guarantee safe exploration
during training, and safe execution at inference time.
A thorough survey of safe RL was presented in~\cite{jmlr:garcia:2015},
which classified techniques into two categories:
those incorporating safety terms in the policy optimality criterion,
and those modifying the exploration process using additional information.
In particular, our work was inspired by~\cite{icml:achiam:2017},
which proposed
a new trust region method,
Constrained Policy Optimization (CPO),
to guarantee policy performance improvement
with near-constraint satisfaction at each iteration,
along with important
results on convergence speed
and worst-case constraint violation.
Still, real-world robot experiments may require exact constraint satisfaction,
though possibly at the cost of less (time and data) efficient training.
Alternatively, constraints can be enforced on the neural network predictions directly:
\textit{a priori},
e.g.,
by placing limits on the range of possible actions before sampling~\cite{icra:chen:2017};
or \textit{a posteriori},
e.g.,
by ensuring that predictions do not send the robot's end effector
outside of a chosen bounding sphere, reprojecting it inside if needed~\cite{icra:gu:2017}.
Our approach extends these ideas by reshaping the range of possible actions
at each time step
to ensure that resulting predictions always satisfy the exact safety constraints
(not approximations thereof),
during both training and inference.
Our work also benefitted from~\cite{icml:amos:2017},
which proposed a neural network architecture to solve QP optimization problems
in the context of SL, with applications to Sudoku puzzles.
We extend this research towards RL
for real-world systems with physical constraints.

\section{Background and Motivation}
\label{sec:problem_statement}

In this section, we summarize the RL algorithm our work builds upon, TRPO (Section~\ref{sec:trpo}).
We then consider the motivating example of a 2D reaching task (Section~\ref{sec:motivating_example})
and illustrate the limitations of RL without explicit consideration of constraints (Section~\ref{sec:preliminary_results}),
namely that learning a safe policy involves unsafe actions in the first place,
with great sensitivity to the associated reward and cost structure.

\subsection{Reinforcement Learning Nomenclature}
\label{sec:trpo}

We consider an infinite-horizon discounted Markov Decision Process (MDP)
characterized by a tuple
$(S, A, P, R, \rho_0, \gamma)$,
with $S$ the set of states (e.g., robot joint configurations),
$A$ the set of actions (e.g., motor commands),
$P: S \times A \times S \rightarrow [0,1]$ the transition probability distribution
to go from one state to another by taking a specific action,
$R: S \times A \times S \rightarrow \mathbb{R}$ the function of associated rewards,
$\rho_0: S \rightarrow [0,1]$ the initial state probability distribution,
$\gamma \in [0, 1)$ a discount factor.
Denoting by
$\pi: S \times A \rightarrow [0,1]$ a stochastic policy,
the objective is typically to maximize the discounted expected return
$\eta(\pi) =
\mathop{\mathbb{E}}\limits_{\tau}\left[
    \sum\limits_{i=0}^{\infty}
    \gamma^iR(\abbstate_i, \abbaction_i, \abbstate_{i+1})
\right],
$
with 
$\tau = (\abbstate_0, \abbaction_0, \dots)$ a state-action trajectory,
$\abbstate_0 \sim \rho_0$,
$\abbaction_i \sim \pi(\cdot | \abbstate_i)$
and $\abbstate_{i+1} \sim P(\cdot | \abbstate_i, \abbaction_i)$.
With the same notations,
we denote by $Q_\pi$ the state-action value function,
$V_\pi$ the value function
and $A_\pi$ the advantage function:
\begin{align}
    Q_{\pi}(\abbstate_i, \abbaction_i) &= \mathop{\mathbb{E}}\limits_{\abbstate_{i+1}, \abbaction_{i+1}, \dots}\left[
    \sum\limits_{l=0}^{\infty}\gamma^l R(\abbstate_{i+l}, \abbaction_{i+l}, \abbstate_{i+l+1})
\right], \\
 V_{\pi}(\abbstate_i) &= \mathop{\mathbb{E}}\limits_{\abbaction_i, \abbstate_{i+1}, \dots}\left[
    \sum\limits_{l=0}^{\infty}\gamma^l R(\abbstate_{i+l}, \abbaction_{i+l}, \abbstate_{i+l+1})
\right], \\
A_{\pi}(\abbstate, \abbaction) &= Q_{\pi}(\abbstate, \abbaction) - V_{\pi}(\abbstate).
\end{align}

We consider in particular the case of a policy $\pi_\theta$ parameterized by
a vector $\theta$ (e.g., neural network parameters).
In TRPO,
the policy $\pi_{\theta_k}$ is iteratively refined by solving the following optimization problem
at each iteration $k$:
\begin{align} \label{eq:trpo}
    &\theta_{k+1} = \mathop{\text{argmax}}\limits_{\theta}
    \mathop{\mathbb{E}}\limits_{\abbstate \sim \rho_{\theta_k}, \abbaction \sim \pi_{\theta_k}} \left[
        \frac{\pi_{\theta}( \abbaction | \abbstate )}{\pi_{\theta_k}(\abbaction|\abbstate)}A_{\pi_{\theta_k}}(\abbstate, \abbaction)
    \right] \\
    &\text{such that} \quad
    \mathop{\mathbb{E}}\limits_{\abbstate \sim \rho_{\theta_k}} \left[
        D_{\text{KL}}(\pi_{\theta_k}(\cdot| \abbstate) || \pi_{\theta}(\cdot | \abbstate) )
    \right]
    \leq \delta_{\text{KL}},
\end{align}
with $\rho_\theta(\abbstate) = \sum\limits_{i=0}^{\infty}\gamma^i P(\abbstate_i = \abbstate)$
the discounted state-visitation frequencies when actions are chosen according to $\pi_\theta$,
$D_{\text{KL}}$ the Kullback-Leibler divergence
(measuring how much two probability distributions differ from each other),
and $\delta_\text{KL}$ a step size
controlling how much the policy is allowed to change at each iteration.
Eq.~\eqref{eq:trpo} is then solved numerically by sampling state-action trajectories
following $\pi_{\theta_k}$ and averaging over samples.
Our work builds on top of the OpenAI Baselines reference implementation for
TRPO\footnote{\url{https://github.com/openai/baselines}},
in which a neural network $\abbnetwork$
predicts state values jointly with actions.
In the rest of the paper,
we refer to it as a procedure
$\abbprocupdatenetwork\left(\abbnetwork, (\abbstate_i, \abbaction_i, r_i, v_i)_{i}\right)$
that iteratively optimizes $\abbnetwork$
from sequences of state-action-reward-value tuples
$(\abbstate_i, \abbaction_i, r_i, v_i)_{i = 0, \dots, N}$
over finite horizon $N$.

\subsection{Motivating Example}
\label{sec:motivating_example}

\begin{figure}[!t]
    \centering
    \subfloat[2-DoF robot, target and obstacle.] {
        \label{fig:2d_reacher:description}
        \begin{overpic}[width=0.5\columnwidth]{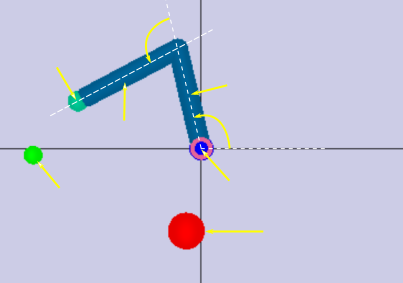}
            \put (67,11) {obstacle}
            \put (7,18) {target}
            \put (18,36) {forearm}
            \put (1,60) {end}
            \put (1,54) {effector}
            \put (58,48) {upper arm}
            \put (20,63) {$\theta_\text{elbow}$}
            \put (57,37) {$\theta_\text{shoulder}$}
        \end{overpic}
    }
    \subfloat[2D-reaching reward structure.] {
        \label{fig:2d_reacher:rewards}
        \begin{tabular}[b]{|c|c|}
            \hline
            \textsc{Type} & \textsc{Reward} \\
            \hline
            $r_\text{dist}$ & $-\abbdistancetargetee$ \\
            \hline
            $r_\text{coll}$ & $-20\abbalphacoll$ \\
             & if collision \\
             & else $0$ \\
            \hline
            $r_\text{prox}$ & $2$ if $\abbdistancetargetee \leq \SI{3}{\centi\meter} $ \\
                             & else $0$ \\
            \hline
        \end{tabular}
    }
    \caption{
        2D reacher description and rewards,
        with
        $\abbdistancetargetee$
        the distance between target and end-effector
        and
        $\abbalphacoll \in \{ 1, 5, 10, 50 \}$ a penalization coefficient.
    }
    \label{fig:2d_reacher}
\end{figure}

We consider a
2-DoF robot with parallel revolute joints.
The robot's base link is rigidly linked to the world
such that it evolves in a plane $(\abbx, \abby)$
perpendicular to the gravity vector $-g\abbz$.
Its two main links, upper arm and forearm, are of length $\SI{0.1}{\meter}$.
The task consists in reaching a target point with the robot's end effector
while avoiding a spherical obstacle, both on the plane $(\abbx, \abby)$.
We denote by $\abbjointpos = (\theta_\text{shoulder}, \theta_\text{elbow})$
the robot's joint angles,
with $\theta_\text{shoulder}$ not limited and $\theta_\text{elbow} \in [-\pi, +\pi]$,
and by $\abbjointvel, \abbjointacc$ its joint velocities and accelerations, respectively.
We depict the 2D reacher environment, $\abbenv_\text{2D}$, in Fig.~\ref{fig:2d_reacher:description}.
Each episode is initialized as follows:
\begin{enumerate}
    \item The robot is reset to the initial state $\abbjointpos = \abbjointvel = \abbjointacc = \abbzeromat$.
    \item The target's coordinates are randomly sampled following a uniform distribution
        of $\pm \SI{0.27}{\centi\meter}$ along $\abbx$ and $\abby$ (possibly out of reach for the robot).
    \item The obstacle is initialized following the same distribution.
        If it collides at initialization with the robot, its position is re-sampled until it is no more the case.
\end{enumerate}
We discretize time into $N=200$ steps of duration $\Delta T = \SI{0.01}{\second}$.
The state vector $\abbstate_i$ at step $i$, or time $t_i = i \Delta T$, is:
\begin{itemize}
    \item $\abbforearmpos$ the 3D position of the forearm's base,
    \item $\theta_\text{elbow}$ the elbow joint angle,
    \item $\abbjointvel$ the joint velocities,
    \item $\abbendeffectorpos$ the 3D position of the end effector,
    \item $\abbtargetpos - \abbendeffectorpos$ the target position $\abbtargetpos$ relative to the end effector,
    \item $\abbobstaclepos - \abbendeffectorpos$ the obstacle position $\abbobstaclepos$, also relative to $\abbendeffectorpos$.
\end{itemize}
The robot is controlled in position by providing,
at step $i$,
the desired joint position at step $i+1$. 
We define a zero-centered action $\abbaction_i$ as
the joint step $\Delta \abbjointpos_i = \abbjointpos_{i+1} - \abbjointpos_{i}$
to perform between two consecutive time steps.
Finally, executing $\abbaction_i$ in the environment yields three rewards,
$r_\text{dist}$ (reward on low distance to target),
$r_\text{coll}$ (collision penalization),
$r_\text{prox}$ (bonus on proximity to target),
detailed in Fig.~\ref{fig:2d_reacher:rewards}.
We denote by $r_i$ the total reward at step $i$.
In order to assess the sensitivity of RL to the reward structure,
we parameterize the collision reward $r_\text{coll}$
with a coefficient $\abbalphacoll \in \{ 1, 5, 10, 50 \}$.

\subsection{Preliminary Results}
\label{sec:preliminary_results}

\begin{figure}[!t]
    \centering
    \includegraphics[width=\columnwidth]{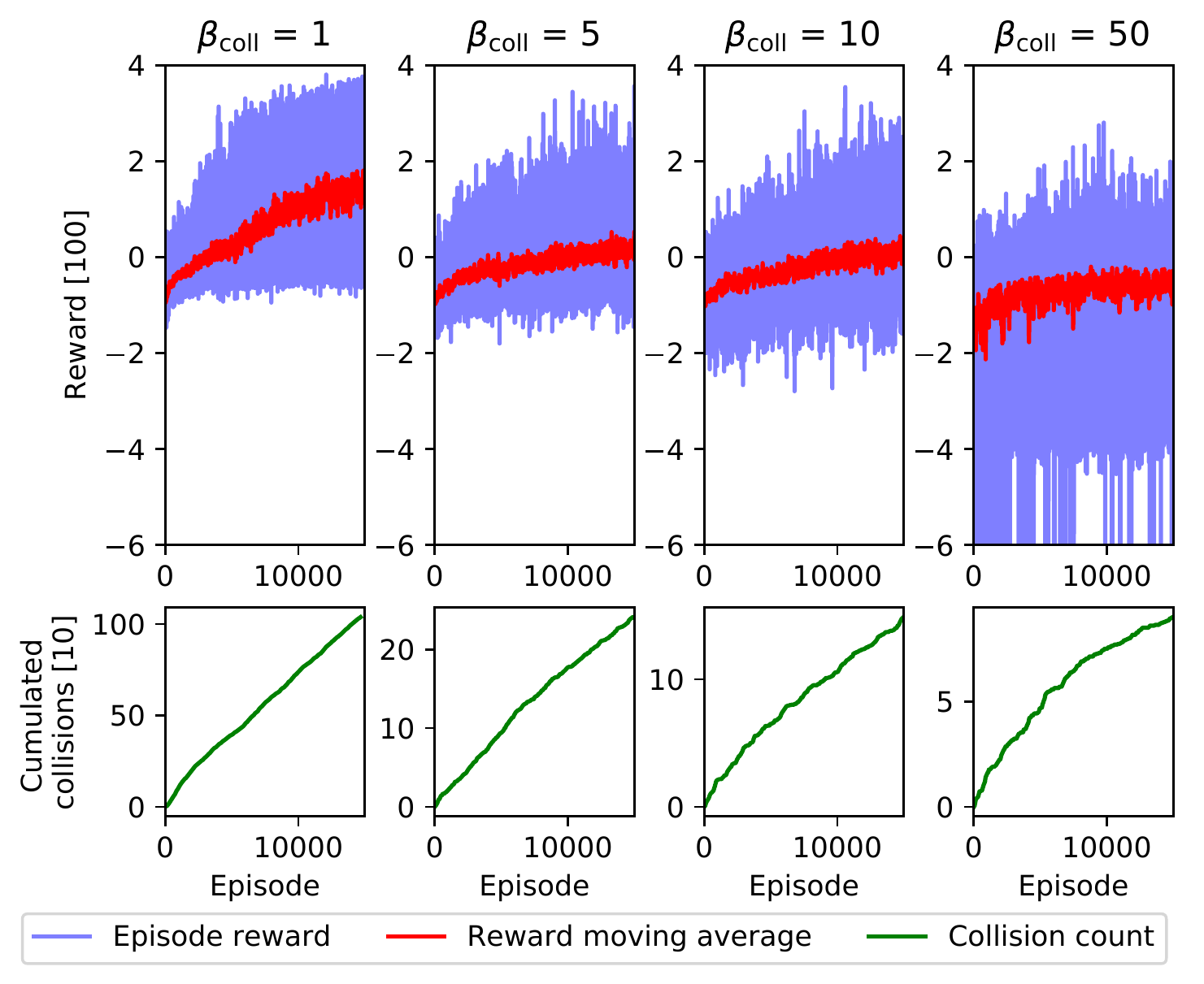}
    \caption{
        Reward and collision count for unconstrained 2D reaching task
        with collision penalization coefficient $\abbalphacoll \in \{1, 5, 10, 50\}$,
        averaged over three seeds.
        Moving average over $40$ consecutive episode rewards.
    }
    \label{fig:safereacherposcosts}
\end{figure}

We implement the environment
$\abbenv_\text{2D}$
within the OpenAI Gym and Roboschool framework~\cite{arxiv:brockman:2016},
which provides a unified interface to train and test neural network policies.
We use the physics engine within Roboschool
to monitor external collisions, between any robot link and the obstacle,
and auto-collisions, between the end effector and the base.
Throughout this paper,
we take as neural network $\abbnetwork$
a simple multilayer perceptron (MLP)
with two hidden layers of size $32$ each.
At each step $i$,
the environment produces a state vector $s_i$,
which is fed into $\abbnetwork$ to produce an action and value pair $a_i, v_i$.
Executing $a_i$ in $\abbenv_\text{2D}$ results in a reward $r_i$ and
an updated state $s_{i+1}$.
We construct such sequences $(s_i, a_i, r_i, v_i)_{i = 0, \dots, N}$
over episodes of up to $N=200$ timesteps, interrupted in case of collision.
On a 8-core computer,
we train $\abbnetwork$
on four instances of $\abbenv_\text{2D}$ in parallel,
with the TRPO procedure
$\abbprocupdatenetwork$
of Section~\ref{sec:trpo}.
We report the evolution of the reward over 15000 episodes of training,
for four values of $\abbalphacoll$,
along with the cumulated number of collisions in Fig.~\ref{fig:safereacherposcosts}.
We observe the following:
\begin{itemize}
    \item The reward evolution throughout training is greatly dependent on the
        collision penalization weight.
    \item With $\abbalphacoll = 1$,
        the low penalty associated to collisions $r_\text{coll} = -20$
        results in collision avoidance never being learnt.
        The optimal strategy consists in quickly going to the target
        to get the proximity reward $r_\text{prox}$
        and possibly hitting the obstacle to start a new episode.
    \item Conversely, with $\abbalphacoll = 50$,
        the large penalty $r_\text{coll} = -1000$
        results in fewer collisions but
        discourages exploration.
        The optimal strategy then consists in staying in the vicinity of the start pose
        and ignoring further targets.
    \item $\abbalphacoll = 5$ and $10$ ($r_\text{coll} = -100$ and $-200$, respectively)
        are comparable in terms of rewards, but the latter results in half the number of collisions.
        This confirms the importance of carefully crafting the reward structure.
\end{itemize}
These results show that, without explicitly consideration of safety constraints during RL,
it is difficult to learn a safe policy without violating such constraints many times.
Overall,
while
we believe it is ultimately possible to design a reward structure that efficiently decreases collisions over time
(e.g., empirically or by inverse reinforcement learning~\cite{iclr:finn:2016}),
for the sake of RL in the real world,
it is crucial to be able to learn to satisfy constraints without ever violating them.

\section{Constrained Optimization Layer}
\label{sec:optlayer}

\subsection{OptLayer Architecture}

\begin{figure}[!t]
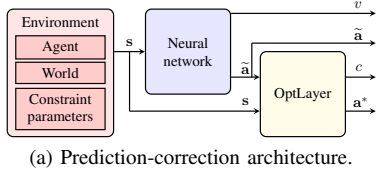
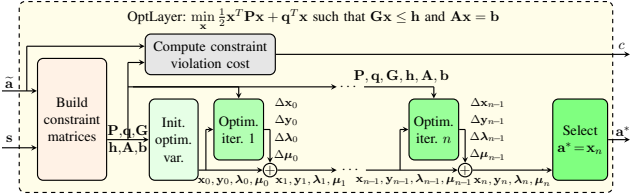

    \centering
    \subfloat[Prediction-correction architecture.] {
        \label{fig:optlayer:overview}
        \resizebox{0.6\columnwidth}{!}{
            \flowchartoverview
        }
    } \\
    \subfloat[
        OptLayer computes the safe action $\abbaction^*$ that is closest to neural network prediction $\abbprediction$,
        while quantifying how much the latter violates safety constraints.
        ] {
        \label{fig:optlayer:details}
        \resizebox{1.0\columnwidth}{!}{
            \flowchartoptlayer
        }
    }
    \caption{
        OptLayer constrained optimization pipeline.
    }
    \label{fig:optlayer}
\end{figure}

With $\abboptvar$ a vector of $\abbnoptvar$ variables to optimize,
we consider quadratic programs (QP) of the form:
\begin{align}
    \min\limits_{\abboptvar} \quad & {
        \frac{1}{2}\abboptvar^T\abbobjmat\abboptvar + \abbobjvec^T\abboptvar
        } \label{eq:qp:obj} \\
    \text{such that} \quad
    & \abbineqmat\abboptvar \leq \abbineqvec \label{eq:qp:ineq} \\
    \text{and} \quad
    & \abbeqmat\abboptvar = \abbeqvec. \label{eq:qp:eq}
\end{align}
Eq.~\eqref{eq:qp:obj} is a quadratic objective in $\abboptvar$,
with $\abbobjmat$ a square matrix and $\abbobjvec$ a vector of respective size $\abbnoptvar \times \abbnoptvar$ and $\abbnoptvar$;
Eq~\eqref{eq:qp:ineq} a set of $\abbnineq$ linear inequalities,
with $\abbineqmat$ and $\abbineqvec$ of respective size $\abbnineq \times \abbnoptvar$ and $\abbnineq$;
and Eq~\eqref{eq:qp:eq} a set of $\abbneq$ linear equalities,
with $\abbeqmat$ and $\abbeqvec$ of respective size $\abbneq \times \abbnoptvar$ and $\abbneq$.
Our goal is to ensure that the action $a_t$ predicted by a neural network $\abbnetwork$
satisfies constraints such as Eqs.~\eqref{eq:qp:ineq} and~\eqref{eq:qp:eq}.
In the following, we denote by $\abbprediction_i$ the action predicted by the neural network $\abbnetwork$
based on the current state $\abbstate_i$.
We propose a constrained optimization layer, OptLayer,
that computes the action $\abbcorrection_i$ that is closest to $\abbprediction_i$
while also satisfying constraints depending on $\abbstate_i$.
We depict the overall architecture in Fig.~\ref{fig:optlayer:overview}.
While multiple solutions are already available for solving QPs alone
(as a forward pass),
end-to-end neural network training requires a backward pass,
in particular computing the gradients of the outputs
with respect to the input parameters.
We thus construct objective and constraint matrices
as differentiable expressions of OptLayer's inputs $\abbstate_i$ and $\abbprediction_i$
(Section~\ref{sec:optlayer:initialization}).
We then solve the QP with an interior point method implemented
as fully differentiable layers
(Section ~\ref{sec:optlayer:solving}).

\subsection{Quadratic Program Initialization}
\label{sec:optlayer:initialization}

Our first goal is to construct objective
and constraint matrices
as differentiable expressions of $\abbprediction_i$ and $\abbstate_i$.
Special care is given to constraint matrices,
which for robotic applications
can be constant or affine in $\abbstate_i$,
require auxiliary state variables,
or activate only under particular conditions.
Although only inequality constraints appear in the following exposition,
equality constraints are treated the same way.

\subsubsection{Objective matrices}
We minimize the (squared) $L^2$ distance between optimization variables $\abboptvar$
and prediction $\abbprediction_i$:
\begin{align}
    \min\limits_{\abboptvar} {
        \frac{1}{2}|| \abboptvar - \abbprediction_i ||^2
    }
    =
    \frac{1}{2}\abboptvar^T \abboptvar
    - \abbprediction_i^T \abboptvar
    + \frac{1}{2}\abbprediction_i^T \abbprediction_i.
    \label{eq:obj:min}
\end{align}
Since $\abbprediction_i^T \abbprediction_i$ does not depend on $\abboptvar$,
it does not affect the optimum.
Eq.~\eqref{eq:obj:min} can thus be written as Eq.~\eqref{eq:qp:obj}, with
$\abbobjmat = \abbidentitymat_{\abbnoptvar}$ the identity matrix of size $\abbnoptvar \times \abbnoptvar$
and $\abbobjvec = -\abbprediction_i$.

\subsubsection{Constant constraint matrices}
\label{sec:optlayer:cst}

In general, the constraints that affect a robot can be static or depend on its state.
For the former, consider for example joint velocity limits:
\begin{align}
    \abbjointvel_\text{min} \leq \abbjointvel \leq \abbjointvel_\text{max},
\end{align}
with $\abbjointvel_\text{min}, \abbjointvel_\text{max}$ velocity limits specified by the manufacturer
(typically, $\abbjointvel_\text{min} = -\abbjointvel_\text{max}$).
At step $i+1$, approximating
joint velocities $\abbjointvel_{i+1}$
with
$\Delta \abbjointpos_{i}/\Delta T$
yields the following inequalities on action hypothesis $\abboptvar = \Delta \abbjointpos_i$,
following Eq.~\eqref{eq:qp:ineq}:
\begin{align} \label{eq:ineq:constant}
    [\abbidentitymat_{\abbnoptvar}] \abboptvar
    \leq
    [\Delta T \abbjointvel_\text{max}]
    \text{ and }
    [- \abbidentitymat_{\abbnoptvar}] \abboptvar
    \leq
    [- \Delta T \abbjointvel_\text{min}].
\end{align}
Eq~\eqref{eq:ineq:constant} thus defines a set of constraint matrices
$\abbineqmatcst = [\abbidentitymat_{\abbnoptvar}, -\abbidentitymat_{\abbnoptvar}]^T$
and
$\abbineqveccst = [\Delta T \abbjointvel_\text{max}, -\Delta T \abbjointvel_\text{min}]^T$
that are constant throughout the robot's motion.

\subsubsection{Constraint matrices affine in the state vector}
Consider now joint position limits $\abbjointpos_\text{min}, \abbjointpos_\text{max}$:
\begin{align}
    \abbjointpos_\text{min} \leq \abbjointpos \leq \abbjointpos_\text{max}.
\end{align}
In this case, the range of possible actions depends on the current joint positions.
Since $\abbjointpos_{i+1} = \abbjointpos_i + \abboptvar$, we have:
\begin{align} \label{eq:ineq:state}
    [\abbidentitymat_{\abbnoptvar}] \abboptvar
    \leq
    [\abbjointpos_\text{max} - \abbjointpos_i]
    \text{ and }
    [- \abbidentitymat_{\abbnoptvar}] \abboptvar
    \leq
    [\abbjointpos_i - \abbjointpos_\text{min}].
\end{align}
Since $\abbjointpos_i$ is a member of the state vector $\abbstate_i$,
we can construct a selection matrix $\abbselectionmatrixlin$
(with only $0$ and $\pm 1$)
such that $[-\abbjointpos_i, \abbjointpos_i]^T = \abbselectionmatrixlin \abbstate_i$.
Eq.~\eqref{eq:ineq:state} defines a set of constraint matrices
that are constant
or affine in $\abbstate_i$:
$\abbineqmatlin = [\abbidentitymat_{\abbnoptvar}, -\abbidentitymat_{\abbnoptvar}]^T$
and
$\abbineqveclin = \abbselectionmatrixlin \abbstate_i + [\abbjointpos_\text{max}, -\abbjointpos_\text{min}]^T$.

\subsubsection{Auxiliary state parameters}
\label{sec:optlayer:aux}

In some cases,
constraint matrices may not be formulated as a simple function
of $\abbstate_i$.
Consider for example limits $\abbjointtorques_\text{min}, \abbjointtorques_\text{max}$
on joint torques $\abbjointtorques$:
\begin{align}
    \abbjointtorques_\text{min} \leq \abbjointtorques \leq \abbjointtorques_\text{max}.
    \label{eq:ineq:jointtorques}
\end{align}
With $\abbgenpos, \abbgenvel, \abbgenacc$ the robot's
generalized coordinates (constant base pose and joint angles $\abbjointpos$),
generalized velocities (zero base velocities $\abbzeromat_6$ and joint velocities $\abbjointvel$)
and
generalized accelerations (zero base accelerations $\abbzeromat_6$ and joint accelerations $\abbjointacc$),
respectively,
let $\abbmassmatrix(\abbgenpos)$ be its mass matrix,
$\abbbiasmatrix(\abbgenpos, \abbgenvel)$ the bias vector comprising Coriolis, centrifugal and gravitational forces,
and $\abbbaseforcetorque$ the 6-element vector of the external forces and torques exerted at the robot's (fixed) base.
In the absence of other contact forces,
the equations of motion are then:
\begin{align}
    \abbmassmatrix(\abbgenpos)\abbgenacc + \abbbiasmatrix(\abbgenpos, \abbgenvel)
    =
    \left[
        \begin{matrix}
            \abbbaseforcetorque \\
            \abbjointtorques
        \end{matrix}
    \right].
    \label{eq:motion}
\end{align}
The first 6 rows of Eq.~\eqref{eq:motion} allow the computation of $\abbbaseforcetorque$.
We take $\abbmassmatrix_{\abbjointtorques}(\abbgenpos)$ and $\abbbiasmatrix_{\abbjointtorques}(\abbgenpos, \abbgenvel)$
as the rows indexed 7 and onwards in 
$\abbmassmatrix(\abbgenpos)$ and $\abbbiasmatrix(\abbgenpos, \abbgenvel)$,
respectively,
such that:
\begin{align}
    \abbjointtorques =
    \abbmassmatrix_{\abbjointtorques}(\abbgenpos) \abbgenacc
    + \abbbiasmatrix_{\abbjointtorques}(\abbgenpos, \abbgenvel),
    \text{ with }
    \abbgenacc = 
    \left[
        \begin{matrix}
            \abbzeromat_6 \\
            \abbjointacc
        \end{matrix}
    \right].
    \label{eq:jointtorques}
\end{align}
At step $i$,
estimating $\abbjointacc_i$ by forward difference
$\abbjointacc_i = (\abbjointvel_{i+1} - \abbjointvel_{i})/\Delta T$
and $\abbjointvel_{i+1} = (\abbjointpos_{i+1} - \abbjointpos_{i})/\Delta T = \abboptvar / \Delta T$
yields:
\begin{align}
    \abbjointacc_i = \frac{\abboptvar}{\Delta T^2} - \frac{\abbjointvel_i}{\Delta T}.
    \label{eq:jointacc}
\end{align}
Combining Eqs.~\eqref{eq:jointtorques} and~\eqref{eq:jointacc}
makes $\abbjointtorques_i$ affine in $\abboptvar$:
\begin{align}
    \abbjointtorques_i =
    \frac{\abbmassmatrix_{\abbjointtorques}(\abbgenpos_i)}{\Delta T ^2}
    \left[
        \begin{matrix}
            \abbzeromat_6 \\
            \abboptvar
        \end{matrix}
    \right]
    + \abbbiasmatrix_{\abbjointtorques}(\abbgenpos, \abbgenvel)
    - \frac{\abbmassmatrix_{\abbjointtorques}(\abbgenpos_i)}{\Delta T}
    \left[
        \begin{matrix}
            \abbzeromat_6 \\
            \abbjointvel_i
        \end{matrix}
    \right]
   .
    \label{eq:jointtorquesinx}
\end{align}
Since 
$\abbmassmatrix_{\abbjointtorques}(\abbgenpos_i)$ and $\abbbiasmatrix_{\abbjointtorques}(\abbgenpos_i, \abbgenvel_i)$
can not be expressed as a trivial function (e.g., linear) of
$\abbstate_i$ in its current form,
we adjust the environment $\abbenv$ to compute them internally
and append them to
$\abbstate_i$.
Explicitly, we compute $\abbmassmatrix_{\abbjointtorques}(\abbgenpos_i)$
and $\abbbiasmatrix_{\abbjointtorques}(\abbgenpos_i, \abbgenvel_i)$
from the robot's joint states and physical parameters
using the KDL library~\cite{misc:smits:2011}
and
store their elements as additional elements of $\abbstate_i$.
The joint torque constraints of Eq.~\eqref{eq:ineq:jointtorques}
can then be expressed
following 
Eq.~\eqref{eq:qp:ineq},
with constraint matrices
$\abbineqmataux$ and $\abbineqvecaux$
assembled
from
$\abbstate_i$ (extended) directly.

\subsubsection{Conditional constraint matrices}
\label{sec:optlayer:cond}
Finally, some constraints may be active only under some conditions.
We consider the collision avoidance constraints of~\cite{rss:kanehiro:2008}.
We denote by $\abbconvexobject_l$ a convex object associated to a robot link $l$,
$\abbclosestpoint_l$ the closest point between
$\abbconvexobject_l$ and the environment,
$\abbdistance_l$ and $\abbdistanceunitvector_l$
the corresponding distance and unit vector, oriented towards $\abbclosestpoint_l$.
If $\abbdistance_l$ is smaller than a threshold $\abbinfluencedistance$, named influence distance,
then the following constraint is enabled:
\begin{align}
    - \xi \frac{d_l - \abbsecuritydistance}{\abbinfluencedistance - \abbsecuritydistance} \leq \abbdistancevel_l,
    \label{eq:velocitydamping}
\end{align}
with $\abbdistancevel_l$ the distance velocity,
$\abbsecuritydistance$ a threshold security distance,
and $\xi$ a positive velocity damping coefficient.
Intuitively, Eq.~\eqref{eq:velocitydamping} ensures that
the distance $d_l$ between the robot and its environment
cannot decrease faster than a chosen rate while $\abbsecuritydistance \leq d_l \leq \abbinfluencedistance$,
and that $d_l$ strictly increases if $d_l < \abbsecuritydistance$.
Eq.~\eqref{eq:velocitydamping} can be equivalently written as a constraint on
$\abbgenvel$,
with $\abbjacobian(\abbgenpos, \abbclosestpoint_l)$
the Jacobian matrix of
$\abbconvexobject_l$ at $\abbclosestpoint_l$
when obstacles are static
(otherwise it suffices to adjust $\abbdistancevel_l$
with their velocities):
\begin{align}
    - \xi \frac{d_l - \abbsecuritydistance}{\abbinfluencedistance - \abbsecuritydistance}
    \leq
    \left[\abbdistanceunitvector_l^T\abbjacobian(\abbgenpos, \abbclosestpoint_l)\right] \abbgenvel.
    \label{eq:velocitydamping:gen}
\end{align}
Similarly to Section~\ref{sec:optlayer:aux},
the state vector $\abbstate_i$ can be extended with the elements of
$\abbjacobian(\abbgenpos, \abbclosestpoint_l)^T \abbdistanceunitvector_l$
to form constraint matrices
$\abbineqmatcond$ and $\abbineqveccond$.
We also extend $\abbstate_i$ with $\abbdistance_l$
and denote by $\abbselectionmatrixtest$ the selection matrix such that
$\abbselectionmatrixtest \abbstate_i = \abbdistance_l$.
With $\abbineqvectest = \left[ \abbinfluencedistance \right] - \abbselectionmatrixtest \abbstate_i$,
conditional constraints and their activation
can be expressed following:
\begin{align}
    \text{If }
    \left\{
        \abbzeromat
        \leq
        \abbineqvectest
    \right\},
    \text{ then enable }
    \left\{
    \abbineqmatcond \abboptvar \leq \abbineqveccond
    \right\}.
    \label{eq:conditional:test}
\end{align}
Note that activation condition 
$\left\{\abbzeromat \leq \abbineqvectest\right\}$
can only depend on state $\abbstate_i$,
the action $\abboptvar$ being optimized subsequently.

\subsection{Quadratic Program Solving}
\label{sec:optlayer:solving}

\subsubsection{Assembling constraint matrices for batch solving}

Using the notations of Section~\ref{sec:optlayer:initialization},
we consider first a set of individual constraint matrices
that are always active:
$\abbineqmatcst, \abbineqveccst, \abbineqmatlin, \abbineqveclin, \abbineqmataux, \abbineqvecaux$.
We concatenate them into base constraint matrices
$\abbineqmatbase = [
    \abbineqmatcst, \abbineqmatlin, \abbineqmataux
]^T$
and
$\abbineqvecbase = [
    \abbineqveccst, \abbineqveclin, \abbineqvecaux
]^T$
to be considered at every step.
When solving only one QP at a time,
conditional constraints matrices
$\abbineqmatcond, \abbineqveccond$
can be appended
to the complete constraint matrices
$\abbineqmat, \abbineqvec$
of Eq.~\eqref{eq:qp:ineq}
only when active, i.e.:
\begin{align}
    \text{If }
    \left\{
        \abbzeromat \leq \abbineqvectest
    \right\}:
    &
    \left\{
        \begin{aligned}
            \abbineqmat &= \left[\abbineqmatbase, \abbineqmatcond \right]^T \\
            \abbineqvec &= \left[\abbineqvecbase, \abbineqveccond \right]^T
        \end{aligned}
    \right.
    ,
    \label{eq:condtrue}
    \\
    \text{otherwise}:
    &
    \left\{
        \begin{aligned}
            \abbineqmat &= \abbineqmatbase \\
            \abbineqvec &= \abbineqvecbase
        \end{aligned}
    \right.
    .
    \label{eq:condfalse:base}
\end{align}
However,
deep neural networks are commonly trained using mini-batches of
multiple input-output pairs in parallel.
A technical consequence of parallel computing
is that all elements processed together must be the same size.
With
$\abbineqmatbasesubset$
(resp. $\abbineqvecbasesubset$)
a
submatrix of $\abbineqmatbase$
(resp. $\abbineqvecbase$)
of size that of $\abbineqmatcond$
(resp. $\abbineqveccond$),
we replace Eq.~\eqref{eq:condfalse:base} with:
\begin{align}
    \text{If not }
    \left\{
        \abbzeromat \leq \abbineqvectest
    \right\}:
    \left\{
        \begin{aligned}
            \abbineqmat &= \left[ \abbineqmatbase, \abbineqmatbasesubset \right]^T \\
            \abbineqvec &= \left[ \abbineqvecbase, \abbineqvecbasesubset \right]^T
        \end{aligned}
    \right.
    .
\end{align}
$\abbineqmat$ and $\abbineqvec$ being of constant size thus enables batch solving.

\subsubsection{Iterative resolution}

Provided objective and constraint matrices
$\abbobjmat, \abbobjvec, \abbineqmat, \abbineqvec, \abbeqmat, \abbeqvec$,
we implement the interior point method for solving QPs described in~\cite{oe:mattingley:2012}.
With $\abbqpx$ still denoting the variables to optimize,
$\abbqps$ slack variables for the QP,
$\abbqpz$ and $\abbqpy$ dual variables respectively associated to the inequality and equality constraints,
finding the optimum $\abbqpx^*$
amounts to solving a sequence of linear systems.
First:
\begin{align}
    \left[
        \begin{matrix}
            \abbobjmat & \abbineqmat^T & \abbeqmat^T \\
            \abbineqmat & -\abbidentitymat_{\abbnineq} & \abbzeromat_{\abbnineq, \abbneq} \\
            \abbeqmat & \abbzeromat_{\abbneq, \abbnineq} & \abbzeromat_{\abbneq, \abbneq}
        \end{matrix}
    \right]
    \left[
        \begin{matrix}
            \abbqpx \\
            \abbqpz \\
            \abbqpy
        \end{matrix}
    \right]
    =
    \left[
        \begin{matrix}
            -\abbobjvec \\
            \abbineqvec \\
            \abbeqvec
        \end{matrix}
    \right]
    .
    \label{eq:qp:initialization}
\end{align}
We then use the solution to Eq.~\eqref{eq:qp:initialization}
to set $\abbqpx_0 = \abbqpx$, $\abbqpy_0 = \abbqpy$,
and initialize $\abbqps_0$ and $\abbqpz_0$ depending on the value of $\abbqpz$.
We thus obtain a starting point $(\abbqpx_0, \abbqps_0, \abbqpz_0, \abbqpy_0)$.
The QP is then solved iteratively.
For
$(\abbqpx_k, \abbqps_k, \abbqpz_k, \abbqpy_k)$
at iteration $k$,
with $\abbqpsmat_k = \mathbf{diag}(\abbqps_k)$ and $\abbqpzmat_k = \mathbf{diag}(\abbqpz_k)$,
update directions are computed by solving linear systems of the form~\cite{oe:mattingley:2012}:
\begin{align}
    &
    \left[
        \begin{matrix}
            \abbobjmat & \abbzeromat_{\abbnineq, \abbnineq} & \abbineqmat^T & \abbeqmat^T \\
            \abbzeromat_{\abbnineq, \abbnoptvar} & \abbqpzmat_k & \abbqpsmat_k & \abbzeromat_{\abbnineq, \abbneq} \\
            \abbineqmat & \abbidentitymat_{\abbnineq} & \abbzeromat_{\abbnineq, \abbnineq} & \abbzeromat_{\abbnineq, \abbneq} \\
            \abbeqmat & \abbzeromat_{\abbneq, \abbnineq} & \abbzeromat_{\abbneq, \abbnineq} & \abbzeromat_{\abbneq, \abbneq}
        \end{matrix}
    \right]
    \!
    \left[
        \begin{matrix}
            \Delta \abbqpx_k \\
            \Delta \abbqps_k \\
            \Delta \abbqpz_k \\
            \Delta \abbqpy_k
        \end{matrix}
    \right]
    \!
    =
    \!
    \abbupdatedirrhs_k,
    \label{eq:qp:update}
\end{align}
with $\abbupdatedirrhs_k$ a vector depending on the current iteration
$(\abbqpx_k, \abbqps_k, \abbqpz_k, \abbqpy_k)$
and QP matrices
$\abbobjmat, \abbobjvec, \abbineqmat, \abbineqvec, \abbeqmat, \abbeqvec$.
\cite{icml:amos:2017}
showed that such problems could be efficiently solved on the GPU
and released a public implementation using the PyTorch framework.
In our custom implementation,
we solved Eqs.~\eqref{eq:qp:initialization} and~\eqref{eq:qp:update}
using linear system solvers provided within the Tensorflow framework.
We iteratively update
$
\abbqpx_{k+1} = \abbqpx_k + \Delta \abbqpx_k,
\abbqps_{k+1} = \abbqps_k + \Delta \abbqps_k,
\abbqpz_{k+1} = \abbqpz_k + \Delta \abbqpz_k,
\abbqpy_{k+1} = \abbqpy_k + \Delta \abbqpy_k,
$
for $k=0, \dots, k_\text{max}$.
While early stopping criteria can be employed for single QP solving,
we take $k_\text{max}$ constant for batch solving.
Noting that the problems we solve typically converge in about $5$ iterations,
we empirically set $k_\text{max} = 10$.
The final value for $\abbqpx$ is then outputted as
the action $\abbaction^*$ closest to the initial prediction $\abbprediction$
that satisfies all safety constraints.
We depict the complete OptLayer pipeline in Fig.~\ref{fig:optlayer:details}.

\section{Constrained Reinforcement Learning}
\label{sec:constrained_reinforcement_learning}

We propose a reward strategy that
ensures that safety constraints are never violated throughout training
and
is readily compatible with
existing RL methods,
such as TRPO.

\subsection{Trajectory Sampling}

\begin{figure}[!t]
    \begin{algorithmic}[1]
        \Procedure{BuildTraj}{$\mathcal{E}, \abbnetwork, \abboptlayer, \abbdoconstrained$}
        \State $S, V, R \gets (), ()$ \Comment{States, values, rewards}
        \State $\widetilde{A}, A^{*} \gets (), ()$ \Comment{Predicted, optimal actions}
        \State $C \gets ()$ \Comment{Constraint violation costs}
        \State $\abbstate = \mathcal{E}$.reset(); $end = \text{false}$ \Comment{Get initial state}
        \While{not $end$}
        \State $S$.add($\abbstate$) \Comment{Store current state}
        \State $\abbprediction, v \gets \abbnetwork(\abbstate)$ \Comment{Predict action, value}
        \label{alg:normal:predictactionvalue}
        \State $\abbaction^{*}, c \gets \abboptlayer(\abbstate, \abbprediction)$ \Comment{Constrain prediction}
        \State
               $\widetilde{A}$.add($\abbprediction$);
               $V$.add($v$);
               $A^{*}$.add($\abbaction^*$);
               $C$.add($c$)
        \If{$\abbdoconstrained$}
            \State $\abbstate, r, end = \mathcal{E}$.do($\abbaction^*$)
            \Comment{Respects constraints}
        \Else
            \State $\abbstate, r, end = \mathcal{E}$.do($\abbprediction$)
            \Comment{Can violate constraints}
        \EndIf
        \State $R$.add($r$) \Comment{Store reward}
        \EndWhile \label{alg:normal:end_while}
        \State \textbf{return} $S, \widetilde{A}, R, V, A^{*}, C$ 
        \EndProcedure
    \end{algorithmic}
    \caption{
        Build a training trajectory for network $\abbnetwork$ and
        optimization layer $\abboptlayer$ in environment $\mathcal{E}$.
        If $\abbdoconstrained = \mathbf{false}$, execute raw predictions.
    }\label{alg:normal}
\end{figure}

Recall from Section~\ref{sec:motivating_example} that for unconstrained RL,
we collect
sequences of state-action-reward-value tuples $(\abbstate_i, \abbprediction_i, r_i, v_i)_{i = 0, \dots, N}$
by iteratively executing neural network predictions $\abbprediction$ in the environment $\abbenv$.
Having established that 
executing unconstrained actions can be dangerous in the real world,
we use OptLayer to produce corrected actions $\abbaction^*$
and execute those instead in $\abbenv$.
Additionally, given constraint matrices
$\abbeqmat, \abbeqvec, \abbineqmat, \abbineqvec$,
it is possible to quantify how much unconstrained predictions $\abbprediction$ violate those.
We define equality and inequality violation costs
$\abbeqviolcost$ and $\abbineqviolcost$, respectively:
\begin{align}
    \abbeqviolcost &= || \abbeqmat \abbprediction - \abbeqvec ||,
    \label{eq:viol:eq} \\
    \abbineqviolcost &= || \max (\abbineqmat \abbprediction - \abbineqvec, \abbzeromat_{\abbnineq}) ||.
    \label{eq:viol:ineq}
\end{align}
From Eq.~\eqref{eq:viol:eq}, $\abbeqviolcost$ increases any time $\abbeqmat \abbprediction \neq \abbeqvec$, element-wise.
Similarly, from Eq~\eqref{eq:viol:ineq}, only elements such that
$\abbineqmat \abbprediction > \abbineqvec$
contribute to $\abbineqviolcost$
by taking the element-wise maximum with zero.
In practice,
the rows of
$\abbeqmat, \abbeqvec$
(resp. $\abbineqmat, \abbineqvec$)
can be scaled with an arbitrary
non-zero (resp. positive)
coefficient
without changing the QP solution.
We thus uniformize their individual contributions in
$\abbeqviolcost$
(resp.  $\abbineqviolcost$)
by normalization with the coefficients making each row of
$\abbeqmat$
(resp. $\abbineqmat$)
of unit norm.
We denote by $\abbviolcost = \abbeqviolcost + \abbineqviolcost$
the total constraint violation
and compute it within OptLayer, together with $\abbaction^*$.
We define a procedure 
$\textsc{BuildTraj}$
in Fig.~\ref{alg:normal},
that takes as inputs
an environment $\abbenv$,
a neural network $\abbnetwork$ with OptLayer $\abboptlayer$,
a flag $\abbdoconstrained$ determining whether constrained or unconstrained actions are executed,
and outputs
states, raw predictions, rewards, values, corrected actions and constraint violation costs
$(\abbstate_i, \abbprediction_i, r_i, v_i, \abbaction_i, \abbviolcost_i)_{i}$.
By setting $\abbdoconstrained = \mathbf{false}$,
unconstrained actions are executed.
We only do so in simulation.
For real-world RL, we set $\abbdoconstrained = \mathbf{true}$ and only execute corrected actions.

\subsection{Policy Update Strategies}

With the procedure $\abbprocupdatenetwork$ described in Section~\ref{sec:trpo},
we update the network $\abbnetwork$
with sequences of the form
$(\abbstate_i, \abbaction_i, r_i, v_i)_{i}$.
We assess four ways to do so.

\begin{enumerate}
    \item{Unconstrained Predictions (UP):}
        We run
        $\abbprocbuildtraj$
        with $\abbdoconstrained = \mathbf{false}$
        and update $\abbnetwork$ with
        $\abbprocupdatenetwork\left(\abbnetwork, (\abbstate_i, \abbprediction_i, r_i, v_i)_{i}\right)$,
        as in Section~\ref{sec:motivating_example}.
        This is our baseline for unconstrained RL.
    \item{Constrained - learn Predictions (CP):}
        We run
        $\abbprocbuildtraj$
        with $\abbdoconstrained = \mathbf{true}$
        and update $\abbnetwork$ with
        $\abbprocupdatenetwork\left(\abbnetwork, (\abbstate_i, \abbprediction_i, r_i, v_i)_{i}\right)$.
        Note that
        different state sequences are produced compared to UP,
        as corrected (not predicted) actions are executed.
    \item{Constrained - learn Corrections (CC):}
        We run
        $\abbprocbuildtraj$
        with $\abbdoconstrained = \mathbf{true}$
        and update $\abbnetwork$ with
        $\abbprocupdatenetwork\left(\abbnetwork, (\abbstate_i, \abbaction^*_i, r_i, v_i)_{i}\right)$.
    \item{Constrained - learn Predictions and Corrections (CPC):}
        We run
        $\abbprocbuildtraj$
        with $\abbdoconstrained = \mathbf{true}$
        and compute discounted rewards $\abbdiscountedreward_i$
        using constraint violation costs $\abbviolcost_i$:
        $\abbdiscountedreward_i = \abbreward_i - \abbviolcost_i$.
        We first update $\abbnetwork$ with
        raw predictions and discounted rewards,
        $\abbprocupdatenetwork\left(\abbnetwork, (\abbstate_i, \abbprediction_i, \abbdiscountedreward_i, v_i)_{i}\right)$.
        We update $\abbnetwork$ a second time,
        using corrected actions and non-discounted rewards from initial sampling:
        $\abbprocupdatenetwork\left(\abbnetwork, (\abbstate_i, \abbaction^*_i, \abbreward_i, v_i)_{i}\right)$.
\end{enumerate}
Intuitively, the CP policy update strategy amounts to considering OptLayer
separately from the network and as part of the environment.
In the CC strategy, we try to learn corrected actions directly,
that can possibly greatly differ from initial neural network predictions.
With CPC, we first associate raw predictions with discounted rewards
before associating corrected actions with better rewards.

\section{Experiments}
\label{sec:experiments}

\begin{figure}[!t]
    \centering
    \subfloat[UR5 robot, obstacle and target.] {
        \label{fig:ur5:description}
        \includegraphics[width=0.49\columnwidth]{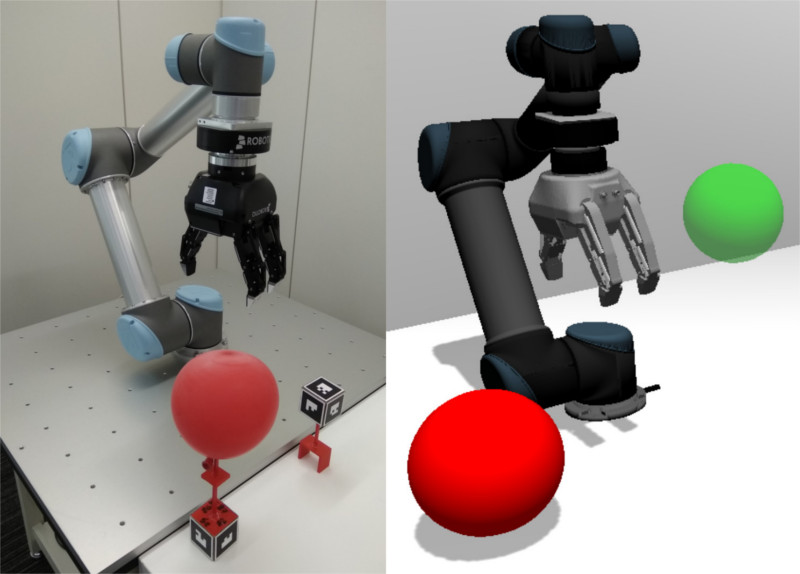}
    }
    \subfloat[3D-reaching reward structure.] {
        \label{fig:ur5:rewards}
        \begin{tabular}[b]{|c|c|}
            \hline
            \textsc{Type} & \textsc{Reward} \\
            \hline
            $r_\text{dist}$ & $-10\abbdistancetargetee$ \\
            \hline
            $r_\text{coll}$ & $-100$ \\
             & if collision \\
             & else $0$ \\
            \hline
            $r_\text{prox}$ & $10$ if $\abbdistancetargetee \leq \SI{5}{\centi\meter} $ \\
                             & else $0$ \\
            \hline
        \end{tabular}
    }
    \caption{
        3D reacher environment,
        with
        $\abbdistancetargetee$
        the distance between target and gripper.
        In simulation, obstacle (red) and target (green) are floating spheres.
        In reality, we position them
        using handles with visual markers.
    }
    \label{fig:ur5}
\end{figure}

We now illustrate the application of OptLayer in the context of RL
to learn 3D reaching tasks using a 6-DoF industrial manipulator (Section~\ref{sec:experiments:environment}),
both in simulation (Section~\ref{sec:experiments:ur5:sim})
and in the real world (Section~\ref{sec:experiments:ur5:real}).

\subsection{3D Reaching with Obstacle Avoidance}
\label{sec:experiments:environment}

We perform our experiments on a 6-DoF industrial manipulator (Universal Robots UR5),
equipped with a
force-torque sensor (Robotiq FT-150)
and a gripper
(Robotiq 3-Finger Adaptive Robot Gripper).
Although we do not use the latter to actually grasp objects in our experiments,
we keep it for the sake of realism,
making it even more crucial to preserve safety.
We control the robot by directly sending desired trajectories through
the Robot Operating System (ROS) middleware,
in joint space,
without making use of any safety mechanism available.
We define a 3D reacher environment, $\abbenv_\text{3D}$,
where the goal is to send consecutive joint commands
to reach a target point in 3D
while avoiding collisions with the environment and the robot itself.
At any step $i$,
joint states $\abbjointpos_i, \abbjointvel_i, \abbjointacc_i$
and actions $\abbaction_i = \abbjointpos_{i+1} - \abbjointpos_i$
are now of size $6$.
The state vector $\abbstate_i$ contains
joint positions and velocities $\abbjointpos_i, \abbjointvel_i$,
end effector position taken at the center of the finger ends $\abbendeffectorposi$,
and target, obstacle 3D positions relative to $\abbendeffectorposi$.
Each episode is run following the same procedure as the 2D case,
except for one adjustment, motivated by real-robot training.
At the end of each episode, we
reset the robot by replaying the trajectory in reverse.
This is important since going back from end to initial pose by joint space interpolation
can lead to collisions in between.
We set the maximum number of time steps to $N=100$,
with duration $\Delta T = \SI{0.1}{\second}$.
We implement $\abbenv_\text{3D}$ to enable parallel training
on multiple physical robots or simulation instances (Gazebo).

\subsection{Evaluating Policy Update Strategies}
\label{sec:experiments:ur5:sim}

To evaluate the UP, CP, CC and CPC policy update strategies,
we initialize safety constraint matrices
as described in Section~\ref{sec:optlayer}.
While we could formulate them manually based on the specifications of our robot system
(UR5 robot, force-torque sensor, gripper),
it is in fact possible to partially automate their definition.
We leverage the availability of standardized robot description formats
(e.g., URDF, MJCF)
often directly released by robot manufacturers.
We thus automatically generate constraint matrices for 
joint limits
by simply parsing
kinematic and physical parameters from such files.
For collision constraints,
we manually construct convex hulls for every robot
link to enable fast distance computation~\cite{icra:pan:2012}.
We then automate the formulation of collision constraint matrices
from a set of distances to monitor,
of the form
$\left\{ (\text{gripper}, \text{forearm}), (\text{elbow}, \text{obstacle}), \dots \right\}$.
In total, we enforce $36$ joint constraints (position, velocity, torque)
and $20$ collision constraints (robot, floating obstacle, environment).

\begin{figure}[!t]
    \centering
    \subfloat[
        Reward and collision count.
        Moving average over $40$ consecutive episodes.
    ] {
        \includegraphics[width=\columnwidth]{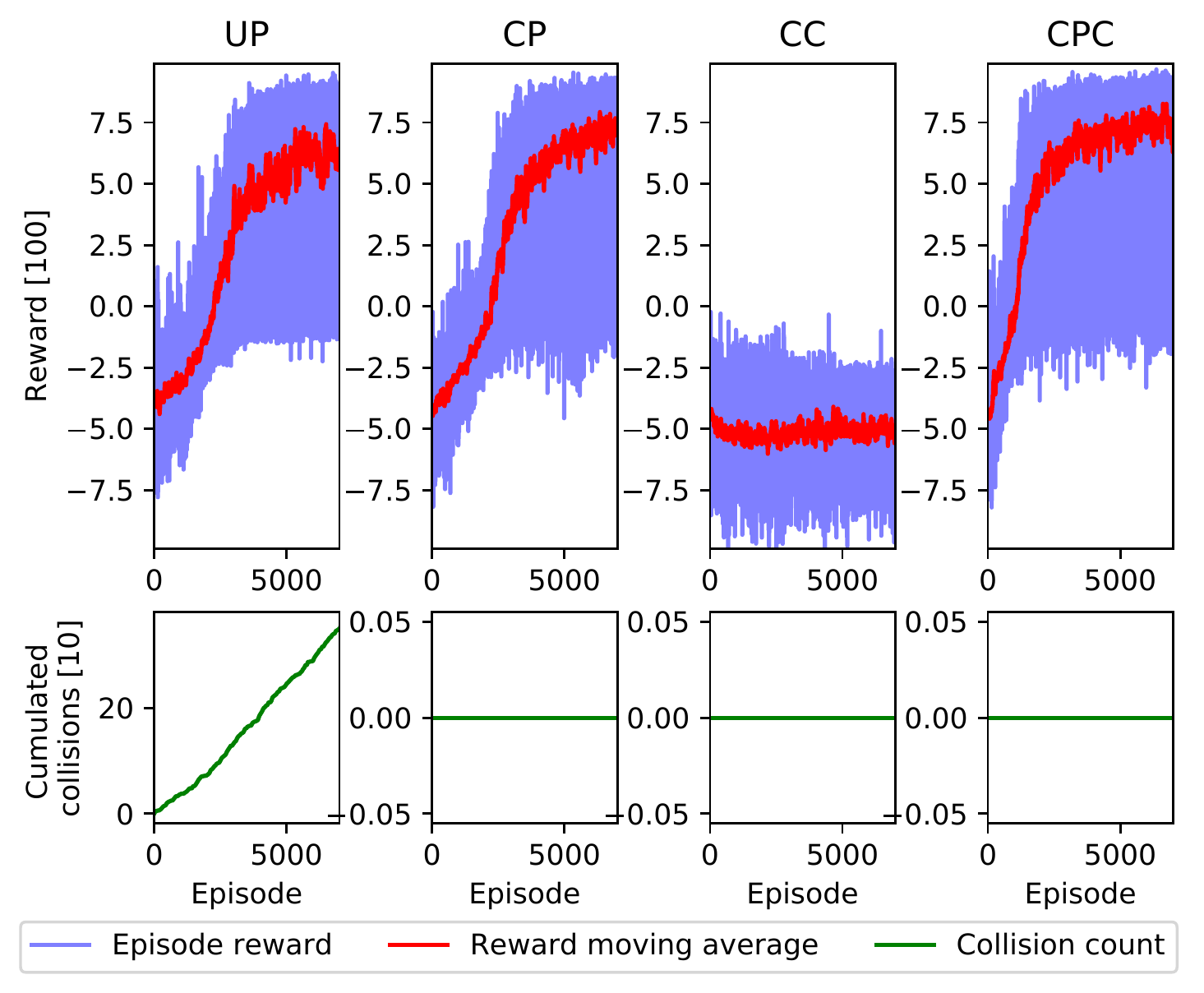}
        \label{fig:safeur5:results:rewards}
    } \\
    \subfloat[Training time over 7000 episodes \textsc{Ep}, steps \textsc{St}, average reward $\textsc{R}\!=\!700$.] {
        \resizebox{1.0\columnwidth}{!}{
            \begin{tabular}[b]{|c|c|c|c|c|c|}
                \cline{2-6}
                \multicolumn{1}{c|}{} & \textsc{Total Time} & \textsc{St/Ep} & \textsc{Time/St} & \textsc{Ep to R} & \textsc{Time to R} \\
                \hline
                UP & \SI{10}{\hour}\SI{19}{\minute} & \SI{88}{} & \SI{60}{\milli\second} & 5350 & \SI{7}{\hour}\SI{53}{\minute} \\
                \hline
                CP & \SI{12}{\hour}\SI{28}{\minute} & \SI{100}{} & \SI{64}{\milli\second} & 4950 & \SI{8}{\hour}\SI{49}{\minute} \\
                \hline
                CC & \SI{12}{\hour}\SI{22}{\minute} & \SI{100}{} & \SI{64}{\milli\second} & N/A & N/A \\
                \hline
                CPC & \SI{12}{\hour}\SI{30}{\minute} & \SI{100}{} & \SI{64}{\milli\second} & 3250 & \SI{5}{\hour}\SI{48}{\minute} \\
                \hline
            \end{tabular}
        }
        \label{fig:safeur5:results:time}
    }
    \caption{
        Training results for 3D reaching with a 6-DoF robot,
        with policy update strategies UP (unconstrained) and CP, CC, CPC (constrained).
    }
    \label{fig:safeur5:results}
\end{figure}

In simulation, we train neural networks on $\abbenv_\text{3D}$
using UP, CP, CC and CPC.
For each policy update strategy, we report rewards and collisions
throughout $7000$ episodes in
Fig.~\ref{fig:safeur5:results:rewards}.
We observe the following.
First, unconstrained training with UC results in the robot successfully learning to reach (top left),
but not in effectively learning collision avoidance (bottom left).
In contrast,
using OptLayer successfully ensures that collisions never happen for all policy update strategies.
In particular, CP realizes rewards that are comparable to UP while never putting the robot at risk.
Recall that in CP,
from the perspective of the network,
OptLayer can be seen as part of the environment,
since the network is updated using raw predictions $\abbprediction$ only.
In constrast, updating the network using only corrected actions $\abbaction^*$ within CC
does not enable learning to reach.
We believe this is due to $\abbaction^*$ being initially too far from $\abbprediction$
to be representative of the stochastic policy we optimize within TRPO,
following Eq.~\eqref{eq:trpo}.
To do so successfully,
CPC
first updates the network with $\abbprediction$ using discounted rewards prior to using $\abbaction^*$.
Subsequently, CPC appears to reach the performance of UP and CP
with far fewer episodes.
For example,
a $700$ average reward is reached in
$5350$ episodes by UP,
$4950$ by CP,
$3250$ by CPC.
Detailed computation times (Fig.~\ref{fig:safeur5:results:time})
show that,
though OptLayer induces some overhead
($\SI{4}{\milli\second}$ per step),
the gain in terms of episodes
makes CPC faster than UP even in terms of time.

\subsection{Real-World Experiments}
\label{sec:experiments:ur5:real}

\begin{figure}[!t]
    \centering
    \includegraphics[height=0.35\columnwidth]{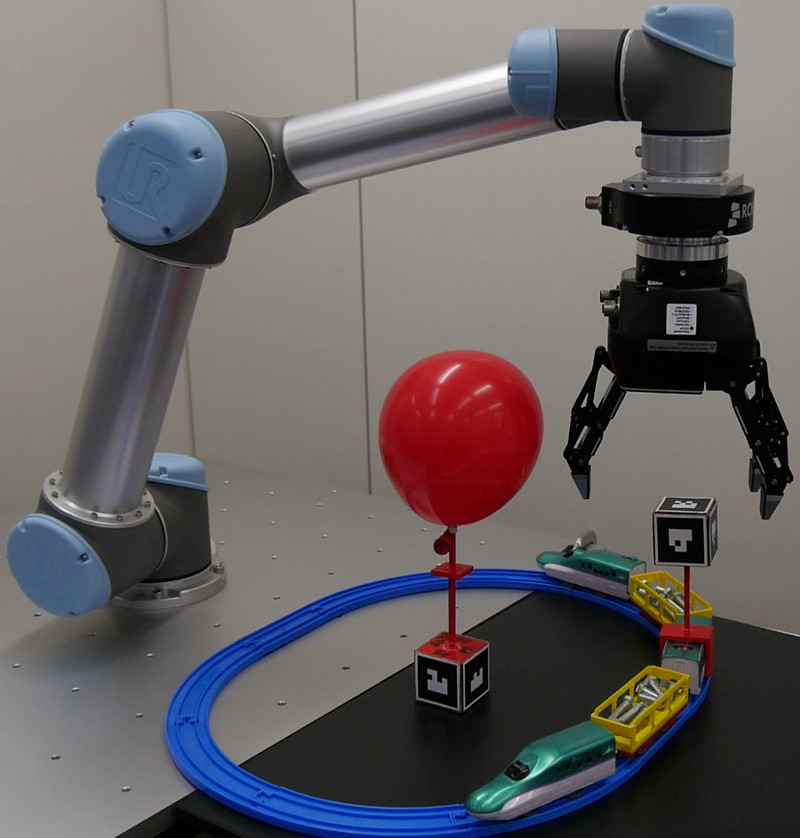}
    \includegraphics[height=0.35\columnwidth]{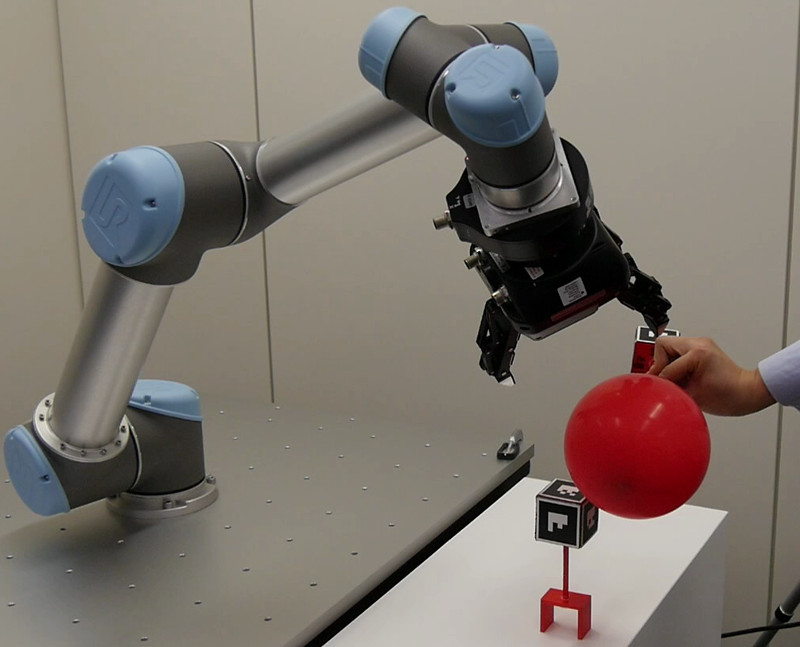}
    \includegraphics[height=0.35\columnwidth]{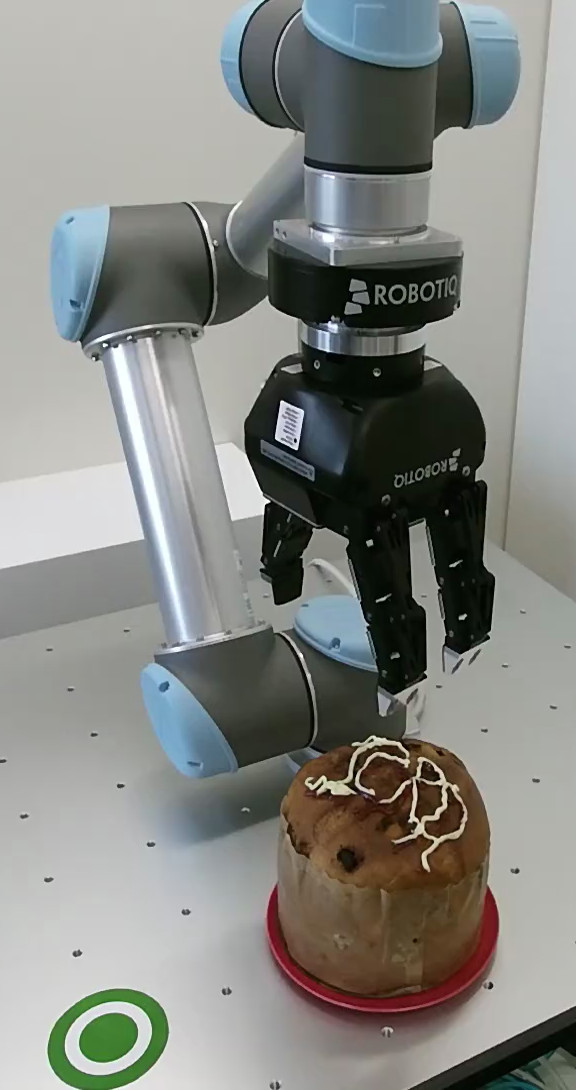}
    \caption{
        From left to right:
        static obstacle, moving target;
        moving obstacle, moving target;
        static obstacle, static target learning on a real robot.
    }
    \label{fig:real}
\end{figure}

Having trained control policies in simulation,
we execute them on a real robot.
In these experiments,
we position obstacle and target with respect to the robot
using physical handles marked with 2D barcodes and tracked
with an RGB camera.
First,
we consider a normal instance of $\abbenv_\text{3D}$
with static target and obstacle.
We execute the unconstrained policy learned with UP, $\abbnetworkup$,
to reach a target behind a balloon (Fig.~\ref{fig:simplereach:un}).
While reaching was successful, the balloon was knocked down,
illustrating that collision avoidance was not effectively learned.
We do not use $\abbnetworkup$ further.
In contrast, using the constrained policy learned with CPC, $\abbnetworkcpc$,
results in successfully reaching the target while avoiding the balloon (Fig.~\ref{fig:simplereach:co}).
As a qualitative example, we consider the case where the target is moving,
carried by a toy train around a balloon (Fig.~\ref{fig:real}, left).
Although $\abbnetworkcpc$ was not explicitly trained on this task,
the target was successfully followed without hitting the balloon.
We then consider the case where the obstacle is also moving.
Through the collision avoidance constraints of Eq.~\eqref{eq:velocitydamping},
we verify that
bringing the obstacle close to the robot forces the latter to move away from it.
These results illustrate that
combining
control policies learned on specific tasks
with OptLayer
can help adapt to changing conditions,
while always maintaining safety.
Finally,
although position control policies can be learned in simulation
for $\abbenv_\text{3D}$,
real-world training can be necessary,
e.g., for tasks involving force control~\cite{iros:inoue:2017}
and physical interaction with the environment~\cite{icra:gu:2017}.
We thus learn a simplified version of $\abbenv_\text{3D}$,
in which the real robot has to reach a static target without touching a
static obstacle.
For ease of visualization, we use a cake for the latter
(Fig.~\ref{fig:real}, right),
never hitting it over $\SI{6}{\hour}$ of training.
Our approach can thus be used to
learn a robot control policy from scratch without active human supervision.
We present our complete experimental results in the supplementary
material\footnote{\url{https://www.youtube.com/watch?v=7liBbk3VjWQ}}.

\section{Discussion and Future Work}
\label{sec:discussion}

Our work establishes that
stochastic control policies
can be efficiently coupled with constrained optimization
to enable RL in the real world,
where safety is crucial.
Our method is simple to implement
and readily compatible with existing RL techniques.
We illustrated its application on neural network policies optimized with TRPO
and demonstrated its effectiveness
on a 3D reaching task with collision avoidance.
Real-world robot experiments showed in particular
that our system could accomodate reasonable changes
in environment and task conditions,
even without dedicated training.

Still, 
our work is subject to some limitations in its current implementation.
While we showed that our approach could speed up training
by better exploiting unsafe predictions,
learning control policies from scratch may remain too time-consuming
even for simple tasks.
We would thus like to investigate novel strategies
for action space exploration tailored to robotics.
Another limitation is that
formalizing safety constraints can be difficult,
e.g., under dynamic model uncertainties.
Even when constraints are well defined,
their activation may be contingent
on the accuracy of other perception systems (e.g., vision).
Our work would thus benefit from further advances in
learning perception and control under uncertainties~\cite{arxiv:christiano:2016,iros:tobin:2017}.
Next,
we intend to tackle
other challenging robot learning and control tasks,
such as bipedal locomotion and multi-agent collaboration.
In the long term,
we plan to extend our approach to problems besides robotics,
such as decision-making systems for finance or healthcare.

\addtolength{\textheight}{-12cm}   



%



\bibliographystyle{IEEEtran}
\bibliography{bib/optlayer}

\end{document}